\documentclass[twoside,11pt]{article}
\usepackage{jair, theapa, rawfonts}

\usepackage{graphicx}
\usepackage{url}
\usepackage{breakurl}
\usepackage[edges]{forest}
\usepackage{color,xcolor}
\usepackage{threeparttable}
\usepackage{booktabs}
\usepackage{multirow}
\usepackage{pifont}
\usepackage{bbding} 
\usepackage{amssymb}
\usepackage{makecell}
\usepackage[figuresright]{rotating}
\usepackage{array}
\usetikzlibrary{trees,positioning,shapes,shadows,arrows}
\tikzset{parent/.style={rounded corners=5pt, align=center, fill=white, draw=red!50!black!90},
    child/.style={align=center, draw=red!50!black!90,text width=3cm,rounded corners=5pt},
    model/.style={align=center, fill=orange!30!yellow!10!, text width=8cm,rounded corners=5pt},
    chars/.style={align=center, text width=6cm,rounded corners=5pt}
}

\ShortHeadings{A Survey on Multi-modal Machine Translation}
{Shen, Shao, Li, Lan, Liu \& Su}
\firstpageno{1}

\begin{document}

\title{A Survey on Multi-modal Machine Translation: Tasks, Methods and Challenges}

\author{\name Huangjun Shen \textmd{(Co-first author)} \email huangjunshen@stu.xmu.edu.cn   \\
       \name Liangying Shao \textmd{(Co-first author)} \email liangyingshao@stu.xmu.edu.cn \\
       \name Wenbo Li \email liwenbo@stu.xmu.edu.cn \\
       \name Zhibin Lan \email lanzhibin@stu.xmu.edu.cn \\
       \name Zhanyu Liu \email zhanyuliu@stu.xmu.edu.cn \\
       \name Jinsong Su \email jssu@xmu.edu.cn \\
       \addr School of Informatics, Xiamen University \\
        361005 Xiamen, China
       }


\maketitle

\begin{abstract}
In recent years, multi-modal machine translation has attracted significant interest in both academia and industry due to its superior performance. It takes both textual and visual modalities as inputs, leveraging visual context to tackle the ambiguities in source texts.
In this paper, we begin by offering an exhaustive overview of 99 prior works, comprehensively summarizing representative studies from the perspectives of dominant models, datasets, and evaluation metrics. Afterwards, we analyze the impact of various factors on model performance and finally discuss the possible research directions for this task in the future. Over time, multi-modal machine translation has developed more types to meet diverse needs. Unlike previous surveys confined to the early stage of multi-modal machine translation, our survey thoroughly concludes these emerging types from different aspects, so as to provide researchers with a better understanding of its current state.
\end{abstract}

\section{Introduction}
\label{Introduction}

As an important natural language processing (NLP) task, machine translation (MT) has undergone several paradigm shifts over the past few decades, from early rule-based translation approaches to current end-to-end neural network based models. However, traditional machine translation models only utilize textual information, neglecting useful information from visual modalities such as images and videos. Therefore, an increasing amount of research focuses on multi-modal machine translation (MMT), which integrates visual information to improve MT. 

\begin{figure*}
\centering
\footnotesize
\includegraphics[width=0.95\textwidth, trim=10 80 0 20,clip]{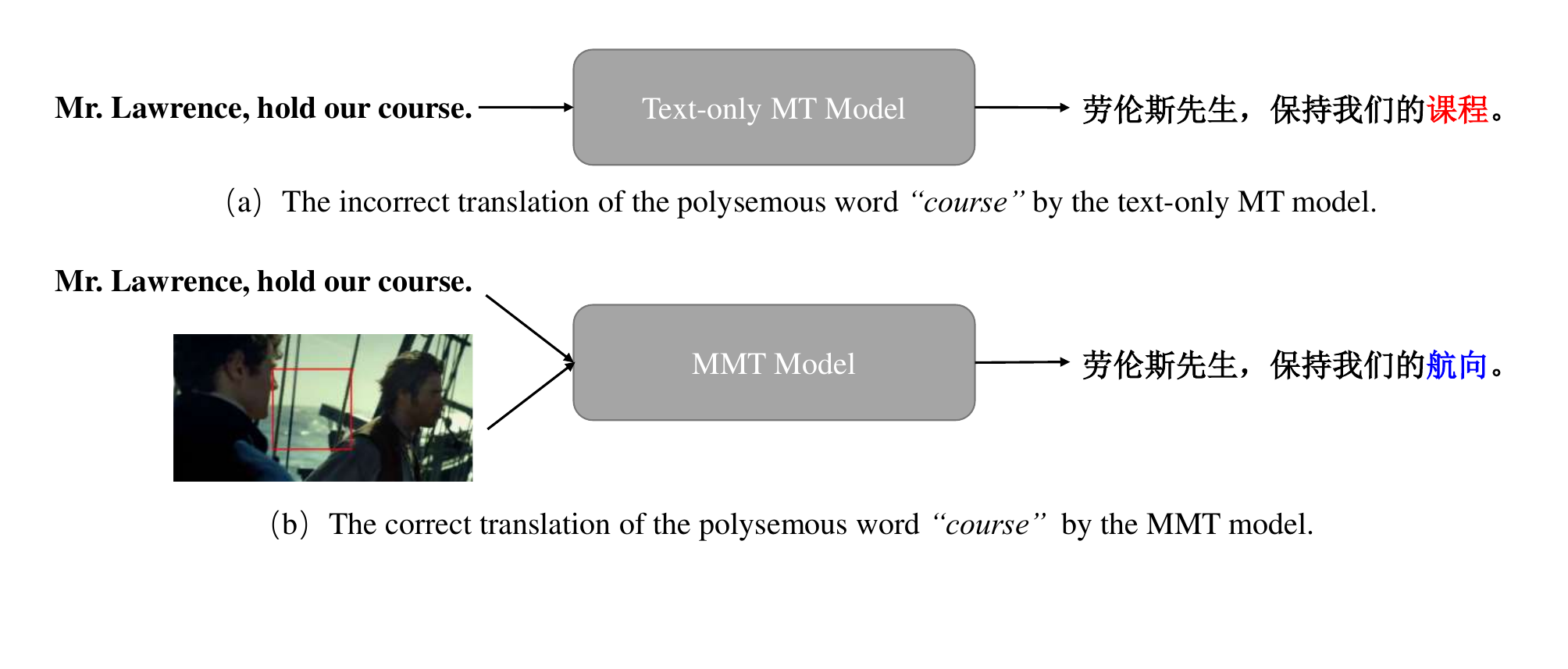}
\setlength{\abovecaptionskip}{5pt}
\caption{An example shows the difference between the conventional text-only MT and MMT models.  }
\label{fig:intro}
\end{figure*}

In general, MMT has important research and application significance. On the one hand, utilizing visual information can provide supplementary context information for source texts, thus alleviating the ambiguity problem caused by the polysemy or omission in text-only MT. Figure \ref{fig:intro} gives an example from the movie \emph{``In the Heart of the Sea''}. It illustrates the difference between MMT and text-only MT. Polysemous words such as \emph{``course''}, which can refer to both a curriculum and a route, pose a significant challenge for text-only MT models. However, with the help of the image indicating the current location at sea, MMT models can easily determine that the word  \emph{``course''} means \emph{``route/direction''}. On the other hand, MMT has been widely used in various applications, such as subtitle translation and cross-border e-commerce product-oriented MT, showing great commercial value. For example, YouTube\footnote{\url{https://www.youtube.com}} is capable of translating subtitles automatically, and Alibaba\footnote{\url{https://www.aliyun.com/product/ai/ecommerce_language}} offers cross-border e-commerce product-oriented MT service.

\begin{figure*}
\centering
\footnotesize
\includegraphics[width=0.9\textwidth, trim=80 110 60 70, clip]{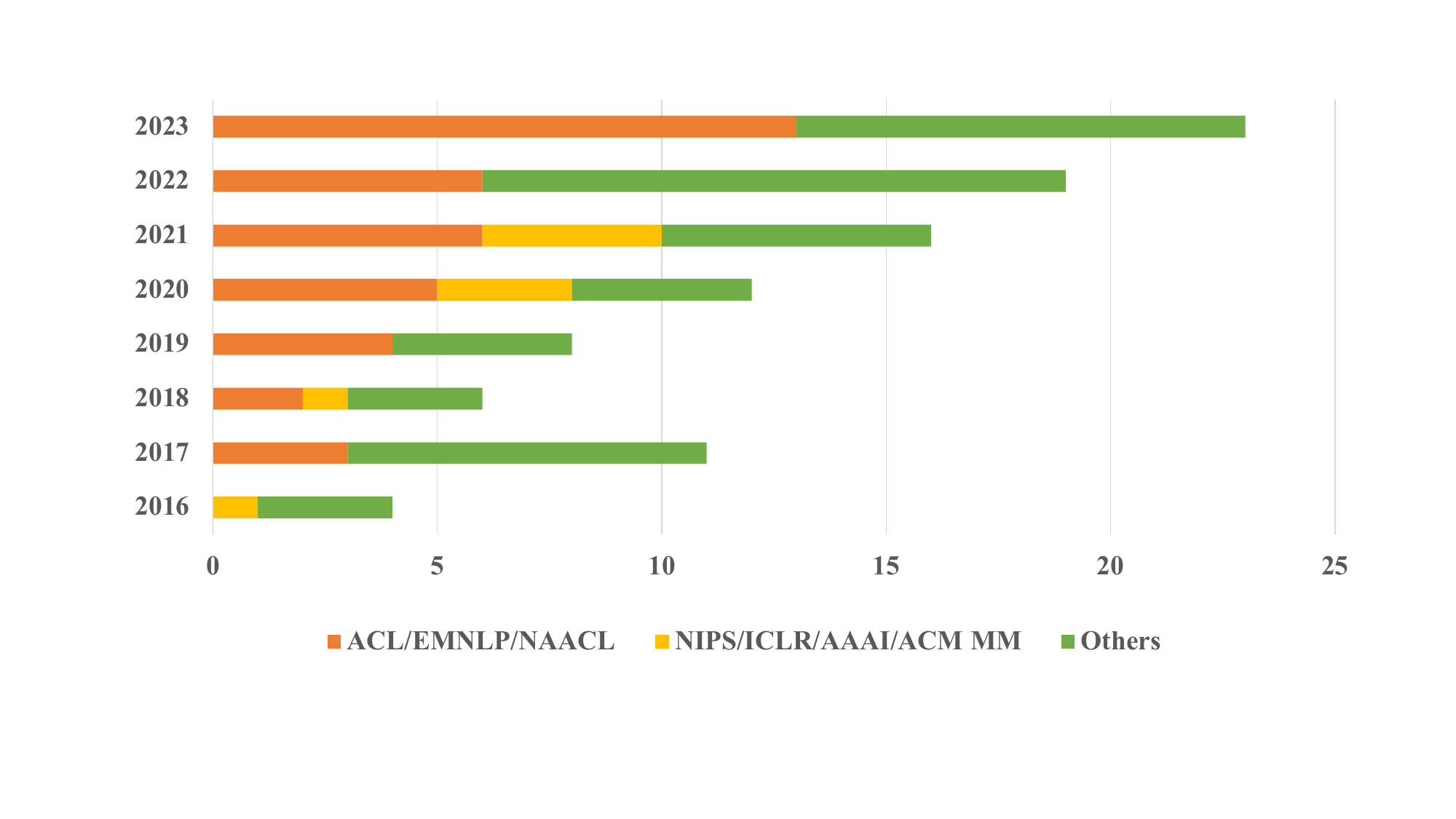}
\setlength{\abovecaptionskip}{5pt}
\caption{Paper publications of MMT at the top computer science conferences and journals.}
\label{fig:paper_cal}
\end{figure*}

Therefore, MMT has attracted much attention and become one of the hot research topics in the community of neural machine translation. Figure \ref{fig:paper_cal} presents the number of papers related to MMT that are published at top computer science conferences and journals. The increasing number demonstrates the growing research passion for this task in recent years.

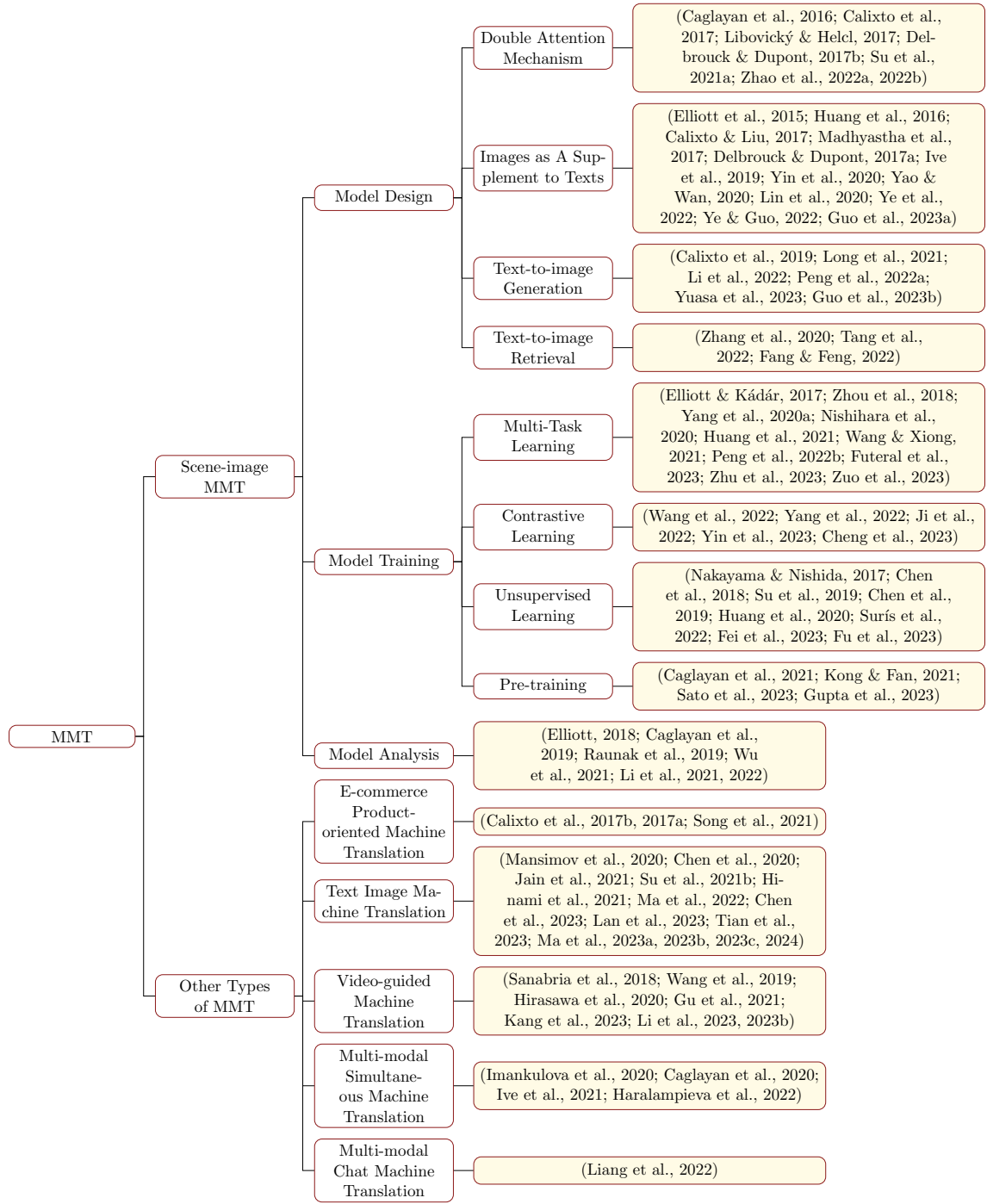
\begin{figure*}[th]
\begin{minipage}[t]{1.0\textwidth}
\resizebox{1.0\textwidth}{!}{
        \begin{forest}
            for tree={
                grow'=east,
                forked edges,
                draw,
                rounded corners,
                node options={},
                text width=2.7cm,
                anchor=west,
            }
            [MMT, parent
                [
                    Scene-image MMT, for tree={child}
                    [ 
                        Model Design,
                        [ 
                            Double Attention Mechanism, [ 
                \shortcite{DBLP:journals/corr/CaglayanBB16,DBLP:conf/acl/CalixtoLC17,DBLP:conf/acl/LibovickyH17,DBLP:journals/corr/DelbrouckD17,DBLP:journals/isci/SuCJZLGWL21,DBLP:journals/ijon/ZhaoKKC22,DBLP:journals/taslp/ZhaoKKC22}
                            , model]
                        ]
                        [ 
                            Images as A Supplement to Texts,
                           [ \shortcite{DBLP:journals/corr/ElliottFH15,DBLP:conf/wmt/HuangLSOD16,DBLP:conf/emnlp/CalixtoL17,DBLP:conf/wmt/MadhyasthaWS17,DBLP:journals/corr/abs-1712-03449,DBLP:conf/acl/IveMS19,DBLP:conf/acl/YinMSZYZL20,DBLP:conf/acl/YaoW20,DBLP:conf/mm/LinMSYYGZL20,DBLP:conf/coling/YeGXT022,DBLP:journals/apin/YeG22,DBLP:journals/taslp/GuoYXY23},model]
                        ]
                        [
                            Text-to-image Generation,
                            [ \shortcite{DBLP:conf/acl/CalixtoRA19,DBLP:conf/naacl/LongWL21,DBLP:conf/cvpr/LiPKCFCV22,DBLP:conf/emnlp/PengZZ22,DBLP:conf/acl/YuasaTKNK23,DBLP:conf/emnlp/GuoF0023},model]
                        ]
                        [
                            Text-to-image Retrieval,
                            [ \shortcite{DBLP:conf/iclr/0001C0USLZ20,DBLP:conf/aclwat/TangZLF22,DBLP:conf/acl/FangF22},model]
                        ]
                    ]
                    [
                        Model Training,
                        [
                            Multi-Task Learning,
                            [ \shortcite{DBLP:conf/ijcnlp/ElliottK17,DBLP:conf/emnlp/ZhouCLY18,DBLP:conf/aaai/YangCZ020,DBLP:conf/coling/NishiharaTNON20,DBLP:conf/emnlp/HuangZZ21,DBLP:conf/aaai/WangX21,DBLP:conf/mir/PengZZ22,DBLP:conf/acl/FuteralSLSB23,DBLP:conf/acl/ZhuSCHWW23,DBLP:conf/emnlp/ZuoLLZXZ23},model]
                        ]
                        [
                            Contrastive Learning,
                            [ \shortcite{DBLP:conf/ijcnn/WangZGYL22,DBLP:conf/emnlp/YangFF22,DBLP:conf/emnlp/JiZZHS22,yin2023multi,DBLP:conf/cikm/ChengZLLZ23},model]
                        ]
                        [
                            Unsupervised Learning,
                            [ \shortcite{DBLP:journals/mt/NakayamaN17,DBLP:conf/aaai/ChenLL18,DBLP:conf/cvpr/SuFBKH19,DBLP:conf/ijcai/ChenJF19,DBLP:conf/acl/HuangHCH20,DBLP:conf/cvpr/SurisEV22,DBLP:conf/acl/0001LZZC23,DBLP:conf/emnlp/FuFHHW0L23},model]
                        ]
                        [
                            Pre-training,
                            [ \shortcite{DBLP:conf/eacl/CaglayanKAMEES21,DBLP:conf/acl/KongF21,DBLP:conf/acl/SatoCS23,DBLP:conf/iccv/GuptaKZ0P023},model]
                        ]
                    ]
                    [
                        Model Analysis,
                            [ \shortcite{DBLP:conf/emnlp/Elliott18,DBLP:conf/naacl/CaglayanMSB19,DBLP:conf/inlg/RaunakCLXM19,DBLP:conf/acl/WuKBLK20,DBLP:conf/emnlp/LiAS21,DBLP:conf/acl/LiLZZXMZ22},model]
                    ]
                ]
                [ Other Types of MMT, for tree={child}
                    [
                        E-commerce Product-oriented Machine Translation,
                        [ \shortcite{DBLP:conf/eacl/MatusovWCSLC17,DBLP:conf/acl-vl/CalixtoSMCW17,DBLP:conf/mm/0003CJLXH21},model]
                    ]
                    [
                       Text Image Machine Translation,
                        [ \shortcite{mansimov-etal-2020-towards,DBLP:conf/icpr/ChenYZYL20,50422,DBLP:conf/icdar/SuLZ21,DBLP:conf/aaai/HinamiIYM21,DBLP:conf/icpr/MaZTHWZ022,DBLP:journals/tmm/ChenYYL23,DBLP:conf/acl/LanYLZ0WHS23,DBLP:conf/emnlp/TianLLGW23,DBLP:conf/emnlp/MaZTZZZ23,DBLP:conf/icdar/MaZTZZZ23a,DBLP:conf/icdar/MaZTZZZ23,DBLP:journals/taslp/MaHWZZZZ24},model]
                    ]
                    [
                        Video-guided Machine Translation,
                        [ \shortcite{DBLP:journals/corr/abs-1811-00347,DBLP:conf/iccv/WangWCLWW19,DBLP:journals/corr/abs-2006-12799,DBLP:conf/acl/GuSCK21,DBLP:conf/acl/KangHP0SCWHS23,DBLP:conf/emnlp/LiSCKL23,DBLP:journals/pami/LiHCHYH23},model]
                    ]
                    [
                        Multi-modal Simultaneous Machine Translation,
                        [ \shortcite{DBLP:conf/wmt/ImankulovaKHK20,DBLP:conf/emnlp/CaglayanIHMBS20,DBLP:conf/eacl/IveLMCMS21,DBLP:journals/jair/HaralampievaCS22},model]
                    ]
                    [ 
                    Multi-modal Chat Machine Translation,
                        [ \shortcite{DBLP:conf/acl/LiangMXCZ22},model]
                    ]
                ]
            ]
        \end{forest}
}
\caption{The Taxonomy of Representative Studies on MMT.}
\label{fig:mmt}
\end{minipage}
\end{figure*}

In this work, we provide a comprehensive review of studies on MMT. Figure \ref{fig:mmt} shows the taxonomy of representative studies. First, we provide a preliminary classification of MMT: scene-image MMT and other types of MMT. For scene-image MMT which current studies mainly focus on, we discuss the relevant studies from three perspectives: model design, model training, and model analysis. As for other types of MMT that have only emerged in recent years, we introduce their task definitions, challenges, and related works in detail. Subsequently, we list the commonly used datasets and evaluation metrics for MMT. Furthermore, we compare the impact of different approaches on model performance, including various image encoding methods and performance-boosting techniques. Finally, we point out the future research directions of this task.

Before our work, few surveys \shortcite{DBLP:journals/mt/SulubacakCGREST20,DBLP:journals/jair/MogadalaKK21} have mentioned MMT and only cover a limited number of related works until 2019, when the task was still in its nascent stage. In contrast, our work fully concentrates on MMT, summarizing up to 99 previous papers and thoroughly including representative studies to date. Moreover, with the development of MMT, more and more types of MMT have been created for different needs. Compared to previous surveys, our work extensively covers these emerging types, providing researchers with a comprehensive understanding of the current state of MMT.


\section{Scene-Image Multi-modal  Machine Translation}
The scene-image MMT task takes source texts and corresponding scene images that usually depict scenes of people's daily activities as inputs. Current studies on MMT mainly focus on this task. In the following subsections, we will introduce these studies from three perspectives: model design, model training, and model analysis.  

\subsection{Model Design}
This line of research concentrates on designing models to utilize the scene image for translation, where the commonly used approaches can be further divided into four categories: using two individual attention mechanisms to extract the text and image contexts (double attention mechanism), using image information as a supplement to textual information, producing text-related image representations and using texts to retrieve images.

\subsubsection{Double Attention Mechanism}
This type of approach considers textual and image information to be equally important, 
and thus uses two attention mechanisms to capture modality-specific contextual information. 
MNMT \shortcite{DBLP:journals/corr/CaglayanBB16} is the first attempt in this aspect. They utilize two bi-directional GRU \shortcite{DBLP:conf/emnlp/ChoMGBBSB14} encoders to encode the source text and image separately, and then apply two individual attention mechanisms to obtain a source and an image context vector, which are finally fused to obtain a multi-modal context vector. Along this line, \citeA{DBLP:conf/acl/CalixtoLC17} introduce an additional gating scalar to quantify the importance of the image context vector at each decoding timestep. Besides, \citeA{DBLP:conf/acl/LibovickyH17} propose a hierarchical attention mechanism for scene-image MMT. At each timestep, the attention mechanism at the bottom generates two modality-specific context vectors, and then the attention mechanism at the top fuses the two context vectors into a final multi-modal context vector.
Unlike prior studies that fuse these context vectors through sum or concatenation operations, \citeA{DBLP:journals/corr/DelbrouckD17} investigate the effectiveness of out product on fusing modality-specific context vectors. To capture fine-grained semantic alignments between text and image, \citeA{DBLP:journals/isci/SuCJZLGWL21} first introduce a bi-directional attention network to refine the textual and image representations, which is the basis of two modality-specific attention mechanisms.
Then, they introduce a co-attention mechanism to better fuse modality-specific context vectors on the decoder side. Instead of relying on global or local spatial image features like previous studies, \citeA{DBLP:journals/ijon/ZhaoKKC22} only attend to semantic image regions to filter the irrelevant information in input images. Based on word-region alignments, \citeA{DBLP:journals/taslp/ZhaoKKC22} employ a word-region similarity matrix to enhance image representations with relevant textual representations.

\subsubsection{Images as A Supplement to Texts}
Some researchers argue that source texts are more critical than input images in scene-image MMT. Thus, they mainly focus on how to leverage images as a supplement to source texts.
The related approaches can be further classified into the following three types.

\textbf{Using Image Information in The Encoder.} In some studies, image information is only incorporated into textual representations during the encoding phase. \citeA{DBLP:conf/wmt/HuangLSOD16} append global and regional image features to the source text, and then feed the concatenated inputs into the encoder for learning contextual representations.  \citeA{DBLP:journals/corr/abs-1712-03449} apply a visual attention mechanism to integrate image information into textual representations. \citeA{DBLP:conf/acl/YaoW20} employ a Transformer with a multi-modal self-attention mechanism to incorporate the information of two modalities. Specifically, they take the concatenation of textual and image representations as query, and retain only textual representations as key and value, which can better extract the relevant image information.
In contrast to the aforementioned methods,\citeA{DBLP:conf/acl/YinMSZYZL20} propose a graph-based multi-modal fusion encoder, which fully exploits fine-grained semantic correspondences between text and image for translation. To build this graph, they treat all words in the source text as textual nodes and the detected image objects as visual nodes. Besides, any two nodes in the same modality are connected by an intra-modal edge, while each textual node representing any noun phrase and the corresponding visual node are connected by an inter-modal edge. Based on the above graph, they sequentially conduct intra-modal and inter-modal fusions in the encoder to update all node states. \citeA{DBLP:journals/apin/YeG22} leverage both global and regional image features to enrich textual representations, and then adopt a multi-modal mixup strategy to fuse textual representations and image features. To narrow the modality representation gap, \citeA{DBLP:journals/taslp/GuoYXY23} propose a layer-level progressive multi-modal fusion strategy. They design a modality difference-aware module to dynamically quantify the modality gap between the source text and image in each encoder layer. Compared to the low-level encoder layers, the high-level encoder layers incorporate more image information into the text.

\textbf{Using Image Information in The Decoder.} Different from the above-mentioned studies, some researchers focus on only learning textual representations during encoding and then introducing image information to assist translation during decoding. For example, \citeA{DBLP:conf/acl/IveMS19} propose a two-stage decoding approach for scene-image MMT. They first only utilize the textual information to generate an initial translation, and then leverage this translation, textual representations and image features together to generate a refined translation. \citeA{DBLP:conf/mm/LinMSYYGZL20} introduce a context-guided multi-modal capsule network to dynamically produce multi-modal context vectors. Concretely, they stack two capsule networks \shortcite{DBLP:conf/nips/SabourFH17} on the last layer of the decoder, which capture the global and regional image features respectively, and then use the timestep-specific source-side context vector as the guiding signal to dynamically produce multi-modal context vectors for translation. To mitigate the noise caused by irrelevant image regions in scene-image MMT, \citeA{DBLP:conf/coling/YeGXT022} compute a mask matrix between each image region and the source text, which selects the most relevant regions to the text for subsequent image features modeling. Finally, they employ a cross-modal gated fusion method to fuse textual and image features. 

\textbf{Using Image Information in Both The Encoder and Decoder.} More researchers concentrate on the effects of simultaneously integrating image information into both the encoder and decoder of scene-image MMT models. \citeA{DBLP:journals/corr/ElliottFH15} study the impacts of initializing encoder and decoder hidden states with global image features. \citeA{DBLP:conf/emnlp/CalixtoL17} explore three ways of incorporating image information: 1) to encode it as the first/last token in the source text, 2) to initialize encoder hidden states, 3) to initialize decoder hidden states. Particularly, they draw the similar conclusion as the previous study \shortcite{DBLP:conf/cvpr/VinyalsTBE15} that using image features to update the decoder hidden states leads to overfitting. 
Instead of using image features to represent image information, \citeA{DBLP:conf/wmt/MadhyasthaWS17} use the predicted class distribution of an image classification network, which contains richer textual semantic information and is more interpretable. Concretely, they initialize the encoder or decoder with the predicted class distribution, and add the projected representation of this distribution to each source word representation.

\subsubsection{Text-to-image Generation}

Some researchers leverage source texts to predict image representations or generate synthetic images to aid the subsequent translation. 
Unlike the approaches mentioned above, no images are required during inference in this line of research. There are two reasons for doing so: 1) imagining visual representations from texts is an instinctive reaction of humans, and these visual representations can act as supplementary context to guide the translation, 2) previous studies typically require the image-text pairs as inputs during inference, however, sometimes it is difficult to acquire such pairs in real-world scenarios.

Typically, \citeA{DBLP:conf/acl/CalixtoRA19} use 
Conditional Variational Autoencoder (CVAE) \shortcite{DBLP:conf/nips/SohnLY15}
to model a joint distribution over translations and images. During training, they generate two multi-modal joint distributions: one is based on the source text, while the other is based on the source text, target text and image. By minimizing the discrepancy between these two distributions, the model can directly generate the multi-modal joint distribution from the source text during inference, from which a multi-modal representation can be sampled to assist the subsequent translation.
Moreover, \citeA{DBLP:conf/naacl/LongWL21} utilize Generative Adversarial Networks (GAN) \shortcite{DBLP:conf/iccv/ZhangXL17} to generate photo-realistic and semantic-consistent image representations conditioned on the source text.
Their training tasks include text-to-image generation, image captioning and scene-image MMT.
\citeA{DBLP:conf/cvpr/LiPKCFCV22} employ an additional visual hallucination Transformer to predict hallucinated image representations. In addition to the conventional translation loss, they introduce a hallucination loss to supervise the model generating
the corresponding hallucinated image representations based on the source text. Furthermore, they propose a consistency loss that narrows the gap between training and inference by drawing close the translation distributions from the hallucinated and ground truth image representations.
\citeA{DBLP:conf/emnlp/PengZZ22} introduce a generator to derive a multi-modal representation from the source text. 
They propose two kinds of knowledge distillation methods to optimize this generator. The first one directs the generator to extract vital image information from the source text, while the second one encourages the generator to profoundly learn the distribution of real images.

Unlike the above-mentioned studies, the following studies utilize synthetic images rather than image representations to aid the translation. Aiming to eliminate the irrelevant content in the image, \citeA{DBLP:conf/acl/YuasaTKNK23} utilize a latent diffusion model \shortcite{DBLP:conf/cvpr/RombachBLEO22} to convert the original image into a synthetic image highly corresponding to the source text, and then perform translation based on the synthetic image.  \citeA{DBLP:conf/emnlp/GuoF0023} concentrate on bridging the gap between real images used in training and synthetic images used in inference. They first utilize a latent diffusion model to generate synthetic images, and then feed real and synthetic images to the translation model respectively. During training, they minimize the gap between two types of images by drawing close their image representations on the source side, and the translation distributions based on two types of images on the target side.

\subsubsection{Text-to-image Retrieval}
Although the studies on text-to-image generation have achieved certain results, they are severely affected by the effects of text-generated images or image representations.
Thus, some researchers resort to image retrieval
to obtain multiple images semantically relevant to the source text, which not only enriches the image representations 
but also extends the applicability of scene-image MMT.

For example, \citeA{DBLP:conf/iclr/0001C0USLZ20} construct a topic-image lookup table by extracting topic words from source texts in the training set and assuming that these words are relevant to the paired images. During retrieval, 
they first search the lookup table with the topic words extracted from the source text to obtain the top-\textit{k} ranked images, and then aggregate the representations of the source text and these retrieved images with a visual attention mechanism.
However, there are three issues in this study: 1) such sentence-level retrieval is difficult to obtain the images that properly match with the source text, 2) the irrelevant image regions also introduce noise information even in the matched images, 3) during inference, the source text may contain out-of-vocabulary (OOV) words that do not exist in the topic-image lookup table. To deal with the first two issues, \citeA{DBLP:conf/acl/FangF22} present a fine-grained phrase-level retrieval approach. Specifically, they extract the grounded image regions related to the noun phrases in the source text to construct a phrase-level image set. Afterwards, they adopt CVAE to reconstruct the noun phrases using the retrieved image regions, which can effectively filter noise image information. To solve the third issue, \citeA{DBLP:conf/aclwat/TangZLF22} use the topic words to retrieve images from a search engine and then employ a text-aware attentive visual encoder to filter noise images.

\subsection{Model Training}
The following studies mainly concentrate on the improvement of model training strategies, where the commonly used strategies include multi-task learning, contrastive learning, unsupervised learning as well as pre-training.

\subsubsection{Multi-task Learning}
Inspired by the success of other NLP tasks, some studies explore multi-task learning to enhance the semantic alignments between text and image.

\citeA{DBLP:conf/ijcnlp/ElliottK17} decompose scene-image MMT into two sub-tasks: text translation and grounded representation prediction. The latter one uses textual representations to predict 
the corresponding image representations.
In this way, the text encoder is encouraged to learn the visually grounded representations for the source language. Via these two sub-tasks, they can fully exploit parallel texts and monolingual image-text data to effectively train their model. Furthermore, \citeA{DBLP:conf/emnlp/ZhouCLY18} propose an image-text shared space learning objective, which draws close the matched textual and image features, while pushing the mismatched ones further. By doing so, the model is able to construct an image-text shared space to generate better modality-shared representations. To fully utilize the image information, \citeA{DBLP:conf/aaai/YangCZ020} propose a visual agreement regularized training loss. They adopt a joint training strategy for both source-to-target and target-to-source translation models, and encourage the models to share the same focus on the image when generating semantic equivalent vision-related words. 
\citeA{DBLP:conf/coling/NishiharaTNON20} supervise the attention mechanism across the source text and image by annotated word-region alignments, and the cross-lingual attention mechanism across the source text and target text is supervised by word alignments between two languages. Inspired by the finding \shortcite{DBLP:conf/naacl/CaglayanMSB19} that entities are most informative in images, \citeA{DBLP:conf/emnlp/HuangZZ21} put forward a reconstruction task to enhance the entity representations. They first replace the embeddings of the visually grounded entities in the source text with the corresponding image object representations to generate a multi-modal input, and then force the model to reconstruct the original source text from the multi-modal input.
\citeA{DBLP:conf/aaai/WangX21} introduce two auxiliary training objectives to assist the main translation task. One is an object-masking loss which grounds the translation on the source-relevant image objects by masking the irrelevant ones. This loss is estimated by the similarity between the masked objects and the source text, which would penalize the undesirable object masking. The other objective is a vision-weighted translation loss that tends to reward the generation of vision-related words. In contrast to previous works that always focus on the alignments of bilingual texts or the combination of the source/target text and the paired image, \citeA{DBLP:conf/mir/PengZZ22} propose a novel framework comprising three tasks to establish a triplet alignment among the source and target texts together with the paired image. The first task is the basic multi-modal translation, the second one utilizes a multi-modal context vector to reconstruct the source text, and the third one employs the multi-modal context vector to perform multi-label classification.
To fully exploit monolingual image-text data, \citeA{DBLP:conf/acl/FuteralSLSB23} introduce an additional image caption denoising task that randomly masks some tokens in the caption and predicts them based on the rest caption and image. Meanwhile, \citeA{DBLP:conf/acl/ZhuSCHWW23} enhance scene-image MMT with an image caption denoising task and a text translation task by utilizing both monolingual image-text data and parallel texts. Their image caption denoising task is similar to the task considered by \citeA{DBLP:conf/acl/FuteralSLSB23}, except that it predicts the complete caption. Besides, they present two ways to incorporate image information: using an image encoder to encode continuous image features, and using an image captioning model to generate keywords of the image, which are then appended to the original source text. 
 \citeA{DBLP:conf/emnlp/ZuoLLZXZ23} propose an auxiliary visual question answering (VQA) task to enhance interactions between two modalities, and utilize large language models (LLMs) to transform traditional datasets like Multi30K \shortcite{DBLP:conf/acl/ElliottFSS16} into VQA patterns.

\subsubsection{Contrastive Learning}
Motivated by the achievements of contrastive learning in other NLP tasks, researchers also implement this strategy to better learn image representations and enhance semantic alignments between text and image.

In this regard, \citeA{DBLP:conf/ijcnn/WangZGYL22} construct positive samples of the original input image via image spatial and appearance transformation, and sample negative samples from the rest images in the same batch. 
Despite the promising performance, scene-image MMT models still face the challenge of input degradation, that is, the models tend to focus more on textual information while overlooking image information. To increase image awareness,  \citeA{DBLP:conf/emnlp/JiZZHS22} employ contrastive learning to maximize the mutual information between text and image on both the source and target sides. On the source side, they use InfoNCE \shortcite{DBLP:conf/iclr/LogeswaranL18} where the matched source texts and images are positive pairs, and the mismatched image-text pairs in the same batch are the negative pairs. On the target side, they follow \citeA{DBLP:conf/emnlp/HuangLPH18} to maximize the discrepancy between two translation distributions based on the original and the deteriorated images, respectively.
To enable zero-shot and few-shot translations for low-resource languages, \citeA{DBLP:conf/emnlp/YangFF22} propose a cross-modal contrastive learning method, which aligns different languages with images as a pivot. They introduce two contrastive learning objectives: 1) the sentence-level objective that involves positive pairs consisting of the matched source texts and images, and negative pairs including other mismatched texts and images in the same batch, 2) the token-level objective that focuses on the source text tokens and the text-aware image tokens. To obtain these text-aware image tokens, they apply an attention mechanism where the query is word-level text tokens, and the key, value are patch-level image tokens, with the text tokens and the generated text-aware image tokens of the same index constituting positive pairs and others constituting negative pairs. 
Likewise, \citeA{DBLP:conf/cikm/ChengZLLZ23} apply contrastive learning at both the sentence and word levels to scene-image MMT. They first leverage an image captioning model to generate the captions of input images, and an object detection model to generate the object labels of image objects. When using contrastive learning at the sentence level, the generated captions are used as positive samples of the original source texts, and other irrelevant source texts are negative samples. When adopting the word-level contrastive learning, the generated object labels are used as positive samples of the corresponding image objects, and other irrelevant object labels in the same image are negative samples. 
\citeA{yin2023multi} extend their previous work \shortcite{DBLP:conf/acl/YinMSZYZL20} by proposing a progressive contrastive learning strategy to refine the model training. Concretely, for each training sample, they simply apply a different dropout mask to the graph encoding to construct a positive sample, and consider two kinds of negative samples: 1) random negative samples that are other multi-modal graphs within the same batch, 2) hard negative samples that are constructed by corrupting the cross-modality alignments or image features of the input graph. Particularly, the hard negative samples are gradually introduced as the training progresses.

\subsubsection{Unsupervised Learning}
In general, with the help of large-scale training corpora, supervised scene-image MMT models show competitive translation performance. However, for many low-resource language pairs, it is expensive to collect large-scale high-quality parallel corpora. Notice that it is easier to obtain the monolingual image-text data for these languages, thus researchers deviate themselves into unsupervised learning based scene-image MMT. 
Moreover, the semantically equivalent visual descriptions in different languages usually refer to the same visual content (e.g., the word \emph{``bicycle''} in English and the word \emph{``vélo''} in French both refer to the bicycle objects in images). Therefore, the utilization of large-scale monolingual image-text datasets can facilitate learning a unified semantic space of different languages with images as a pivot.

\citeA{DBLP:journals/mt/NakayamaN17} first explore zero-resource scene-image MMT. They apply a pair-wise ranking loss to align the matched image-text pairs in both source and target languages, which can force encoders to map different modalities into a common multi-modal semantic space. Besides, the decoder is required to generate the target language description of the input image. 
\citeA{DBLP:conf/aaai/ChenLL18} divide image captioning into two steps: the image is first translated to a source sentence using an image-to-source captioning model, and then the source sentence is translated to a target sentence using a source-to-target translation model. Notice that only the source-to-target translation model is utilized during inference. Different from \shortcite{DBLP:conf/aaai/ChenLL18}, \citeA{DBLP:conf/cvpr/SuFBKH19} introduce two training objectives, the denoising auto-encoding loss based on the source text and image, and the cycle-consistency loss based on back-translation \shortcite{DBLP:conf/acl/SennrichHB16} which pulls close the input text and the output of the back-translation model with the image as a pivot. Besides, they also design a controllable attention module to deal with both uni-modal and multi-modal inputs. \citeA{DBLP:conf/ijcai/ChenJF19} propose to learn the translation in an easy-to-hard progressive way. Their model first learns the word-level translation by generating image captions in the source and target languages, and then generates source-target pseudo text pairs pivoted on the same images. Afterwards, the model learns the sentence-level translation by re-weighting the pseudo pairs at both the sentence and token levels. Besides, \citeA{DBLP:conf/acl/HuangHCH20} define four tasks for joint training. The first one is an unsupervised scene-image MMT task based on back-translation, and the second task performs image-text matching in both source and target languages to map two modalities into a modality-shared semantic space. The third task feeds the image into pre-trained image captioning models of different languages, generating pseudo text pairs pivoted on the same image for translation. Similar to the first one, the last task pulls close the texts generated by image captioning models and the outputs of the back-translation models. \citeA{DBLP:conf/cvpr/SurisEV22} introduce a transitive relation to estimate the similarity between two samples in the monolingual image-text data: if two images are similar to each other, their corresponding texts are also similar. Based on this, they propose several contrastive learning objectives to strengthen the semantic alignments between text and image, including image-to-image, image-to-text, and text-to-text contrastive objectives. \citeA{DBLP:conf/acl/0001LZZC23} leverage language scene graphs (LSGs) and visual scene graphs (VSGs) to better represent source texts and images. They design four learning strategies for unsupervised training. The first one is cross-modal scene graph (SG) aligning, which encourages the textual and visual nodes that serve a similar role in the LSG and the VSG to be closer. The second strategy aims to reconstruct the source text from the VSG, and the image representations from the LSG. The third one performs back-translation with the SG as a pivot, and the last strategy draws close the texts generated by image captioning models and the outputs of the back-translation models.
\citeA{DBLP:conf/emnlp/FuFHHW0L23} incorporate images at the word level to augment the lexical mappings of different languages. They concatenate the features of related images to the embeddings of corresponding words in the source text, and modify the embedding layer information. Based on this multi-modal input, they train the model with unsupervised back-translation and denoising auto-encoding. Besides, a mask matrix is adopted to highlight the relationship between images and their corresponding subwords and isolate the impact of images on other words.

\subsubsection{Pre-traning}
In recent years, the pre-training and fine-tuning paradigm has gained wide attention in both academia and industry due to its superior performance. This paradigm enables scene-image MMT models to acquire foundational knowledge through pre-training tasks, and then enhance their performance on downstream tasks via fine-tuning. In scene-image MMT, tasks related to cross-modal alignments between text and image are often introduced during pre-training, which allows models to generate better modality representations for downstream applications. 

In this respect, \citeA{DBLP:conf/eacl/CaglayanKAMEES21} present a pre-training task named visual translation language modeling. Specifically, the model takes the source text, target text and the detected image objects as inputs. Subsequently, a portion of the input tokens are randomly masked, and the model is required to predict the masked tokens. If an image object is masked, the model is required to restore its correct object label. 
Moreover, \citeA{DBLP:conf/acl/KongF21} utilize a large-scale external image-text dataset as pre-training data, and feed their model with the source text, detected image objects and their object labels as inputs. They introduce two pre-training tasks: the first task aims to restore the text tokens masked in the inputs, and the second one is to judge whether the given text and image are matched.
\citeA{DBLP:conf/acl/SatoCS23} point out that the common random selection strategy adopted in masked language modeling ignores the fact that task-related information varies from token to token. Therefore, they design a more informed masking strategy which masks more pronouns and objects with gender information.
To mitigate the scarcity of annotated scene-image MMT data, especially for low-resource languages, \citeA{DBLP:conf/iccv/GuptaKZ0P023} propose a two-stage learning approach with the pre-trained mBART \shortcite{DBLP:journals/tacl/LiuGGLEGLZ20} as the text model and M-CLIP \shortcite{DBLP:conf/acl/ChenHCDSJLPW23} as the image encoder. In the first stage, the model is trained with the image captioning task by using M-CLIP to encode the image, which forces the decoder to rely on image information to generate corresponding captions.
In the second stage, the model is trained with the text-to-text translation task by using M-CLIP to obtain relevant image information from the source text.

\subsection{Model Analysis}

The basic intuition behind scene-image MMT is that models can improve translation quality by incorporating image context. However, some studies show that the utilization of image context does not consistently enhance model performance. In order to better understand the role of image context in scene-image MMT, researchers have conducted a series of analyses, which suggest that image context can play an important role when textual context is ambiguous or insufficient, but tends to be less effective when textual context is sufficient. 

To determine whether scene-image MMT models are aware of image context, \citeA{DBLP:conf/emnlp/Elliott18} quantify the performance difference of a model given the congruent image or a random incongruent image. They evaluate some publicly available models and find that not all models actually use the image context to produce better translations. However, \citeA{DBLP:conf/naacl/CaglayanMSB19} posit that the texts in the widely-used Multi30K dataset are often very simple, short and repetitive, leading to the limited effectiveness of image context. To investigate this hypothesis, they introduce several input degradation strategies that remove some crucial information from source texts to create limited textual context. The results indicate that models are capable of leveraging the input images to generate better translations under such limited textual context. Furthermore, \citeA{DBLP:conf/inlg/RaunakCLXM19} follow \citeA{DBLP:conf/naacl/CaglayanMSB19} to reconduct experiments on the large-scale How2 dataset \shortcite{DBLP:journals/corr/abs-1811-00347}, where the texts are longer and non-repetitive. Their findings reveal that the quality of visual embeddings rather than the complexity of texts in the existing datasets should be improved. Along this line, \citeA{DBLP:conf/acl/LiLZZXMZ22} explore the impact of image encoders of scene-image MMT models. As implemented in \shortcite{DBLP:conf/naacl/CaglayanMSB19}, they design some probing tasks and find that stronger image encoders are more helpful for learning translation from the image modality. Meanwhile, they also find that models with the enhanced image features achieve improvements in the BLEU score \shortcite{DBLP:conf/acl/PapineniRWZ02} but show no advantage in the probing tasks, which indicates current automatic evaluation metrics might not be suitable for scene-image MMT. To deeply analyze the contribution of image context, \citeA{DBLP:conf/acl/WuKBLK20} adopt the gated fusion approach during training, which allows the model to voluntarily decide the usefulness of image context. Through the gating matrix, they find that image context only influences the early training stage when textual representations are poorly learned. They further discover that under sufficient textual context, the improvements achieved by the multi-modal models over text-only models result from the regularization effect of image context. Using gender-specific images, \citeA{DBLP:conf/emnlp/LiAS21} explore the translation from a gender-neutral language into a language with natural gender. The results show that the integration of image context largely assists models in inferring the correct gender.

\section{Other Types of Multi-modal Machine Translation}
Although the majority of research has primarily centered around scene-image MMT, there are also notable efforts dedicated to other types of MMT, including e-commerce product-oriented MT, text image machine translation (TIMT), video-guided MT, multi-modal simultaneous machine translation (multi-modal SiMT) and multi-modal chat MT. Each of these tasks exhibits distinct characteristics, and we will introduce them from three aspects: task definition, challenge and related works.

\subsection{E-commerce Product-oriented Machine Translation}
In this task, the translation models take both product images and their descriptions as inputs, and then output description translations. It finds wide applications in cross-border e-commerce, providing significant convenience for consumers interested in purchasing goods from overseas. 
Overall, the research on this task still faces the following challenges. First, this task suffers from data scarcity. Second, product descriptions often contain specialized jargons that are ambiguous to be translated without product images.
Third, unlike conventional image descriptions, product descriptions are related to images in more complex ways, involving various visual aspects such as objects, shapes, colors, or even subjective styles.

\shortcite{DBLP:conf/eacl/MatusovWCSLC17} is the first attempt to explore this task. First, the authors build a bilingual product description dataset collected from an online shop website. In this dataset, each sample is made up of a product image, an English description and its German translation. Then they compare the performance of three different models on this dataset: a phrase-based statistical MT (PBSMT) model, a text-only MT model and a MMT model. Their findings reveal that the PBSMT model performs best, followed by the MMT model, and the text-only MT model performs worst.  In addition, they observe that using the MMT model to rerank the outputs of the PBSMT model can significantly improve the Translation Edit Rate (TER) score \shortcite{DBLP:conf/amta/SnoverDSMM06}. In order to further evaluate the performance of different translation models, \citeA{DBLP:conf/acl-vl/CalixtoSMCW17} conduct a human evaluation on the outputs of the aforementioned three models. The results indicate that human evaluators prefer PBSMT translations to both the text-only MT and MMT models in over 56\% of the cases. Nonetheless, human evaluators rank the translations from the MMT model higher than those from the text-only MT model in over 88\% of the cases, suggesting that images indeed assist translation. 
Compared to \shortcite{DBLP:conf/eacl/MatusovWCSLC17}, \citeA{DBLP:conf/mm/0003CJLXH21} construct a larger and more complex dataset, where each sample consists of an English product description, its Chinese translation, product images of different colors and poses, along with product categories and attribute labels. To learn better semantic alignments between bilingual texts and product images, they propose a unified pre-training and fine-tuning framework for this task. Concretely, they introduce three pre-training tasks. The first task aims to reconstruct masked words in both languages with bilingual textual and image context information, the second task conducts semantic matching between the source text and image, and the third task masks the source words conveying product attributes, forcing the model to predict the attributes according to the input product images.

\subsection{Text Image Machine Translation}
In this task, images containing texts in the source language are fed into the translation models to generate either the translations in the target language or images containing the translations. There are a number of commercial applications involving TIMT, such as Google Translate's Instant Camera\footnote{\url{https://blog.google/products/translate/googletranslates-instant-camera-translation-gets-upgrade}} and Google Lens\footnote{\url{https: //ai.googleblog.com/2019/09/giving- lens- new- readingcapabilities-in.html}}.
In sum, the research on this task faces the following challenges. First, data scarcity still remains a major challenge for this task. Second, dominant approaches are cascaded, suffering from error propagation and high latency. They first use an optical character recognition (OCR) model to detect source texts in the input image, followed by a text-only MT model to translate these texts. Sometimes, the translations need to be rendered back into the original input image with optimized font size and location. In this way, existing large-scale OCR and MT datasets and various submodels can be exploited to construct high-quality cascaded models.

To address the above challenges, several studies develop their own datasets, such as BLATID \shortcite{DBLP:journals/tmm/ChenYYL23} and OCRMT30K \shortcite{DBLP:conf/acl/LanYLZ0WHS23}, and present various end-to-end frameworks for this task.
In this regard, \citeA{mansimov-etal-2020-towards} explore generating images containing translations for the first time. During training, their model predicts the next timestep target image conditioned on the source image and ground truth target image at the previous timestep.
Along this line, \citeA{DBLP:conf/emnlp/TianLLGW23} regard the source and target images as pixel sequences converting from their grayscale maps, and apply Byte Pair Encoding (BPE) \shortcite{DBLP:conf/acl/SennrichHB16a} to segment the images.
Different from the first two studies, the following studies mainly focus on generating target texts rather than target images. For example, \citeA{DBLP:conf/icpr/ChenYZYL20} adopt a multi-task learning framework with two sub-tasks. The main task tends to translate the source image into the target text, while the auxiliary task aims to recognize the source text from the source image. 
Moreover, \citeA{DBLP:conf/emnlp/MaZTZZZ23} incorporate the recognition history textual information into the translation decoder via the attention mechanisms when optimizing the model with the above two sub-tasks. 
Unlike the above-mentioned studies with two sub-tasks, 
\citeA{DBLP:conf/icpr/MaZTHWZ022} improve the model performance by jointly training with OCR, MT and TIMT tasks. In this way, the model can fully exploit large-scale external OCR and MT datasets to enhance its image encoder and target text decoder. In \shortcite{50422}, the translation decoder is initialized with a text decoder pre-trained on MT data. Experimental results indicate that their model outperforms the cascaded model in the scenarios of both single-line and multi-line translations. Furthermore, \citeA{DBLP:conf/icdar/SuLZ21} combine pre-training with multi-task learning methods. In the first stage, they pre-train an OCR model to recognize the source text from the image and a MT model to translate the source text. Then in the second stage, the OCR encoder and text-only MT decoder, along with an additional module to bridge the semantic gap between the encoder and decoder, are integrated to construct an end-to-end model, which performs both OCR and TIMT tasks. 
\citeA{DBLP:journals/tmm/ChenYYL23} present a multi-hierarchy cross-modal knowledge distillation strategy. They first pre-train a teacher model with a large bilingual text corpus. Subsequently, they conduct the global knowledge distillation by pulling close the outputs between the teacher encoder and the student encoder, while the local knowledge distillation is performed by elementwise matching the representations derived from the cross-attention mechanisms of the teacher and student models. Furthermore, \citeA{DBLP:conf/icdar/MaZTZZZ23} introduce a multi-teacher knowledge distillation approach, which can transfer different kinds of knowledge to corresponding sub-modules of a student model. First, they pre-train an OCR model and a text-only MT model as the teacher models. Afterwards, the student image encoder is optimized with the guidance from the OCR image encoder, and the student contextual encoder and decoder are improved by transferring the knowledge from the text-only MT encoder and decoder. Particularly, both sentence-level and token-level knowledge distillation are incorporated to better enhance the translation performance. The previous study \shortcite{DBLP:conf/icdar/SuLZ21} links the OCR encoder and MT decoder to facilitate the TIMT task, which, however, ignores the gap between the OCR and MT tasks and thus leads to limited performance. To deal with this issue, \citeA{DBLP:conf/icdar/MaZTZZZ23a} propose an architecture with two types of adapters to eliminate the task gap. The first type is inserted between the OCR image encoder and the text-only MT encoder, aiming to align the image embeddings and text embeddings. The second type connects the OCR contextual encoder and the text-only MT decoder, with the aim to align the contextual semantic feature space. 
To improve model performance, \citeA{DBLP:journals/taslp/MaHWZZZZ24} adopt both intra-modal and inter-modal contrastive learning. When using the intra-modal contrastive learning, they mainly consider two kinds of representation differences: 1) text-text, where the source texts and the outputs of the back-translation models are positive pairs, 2) image-image, where different-format images corresponding to the same source texts are positive pairs. Besides, the inter-modal contrastive learning draws close the representations of the matched source texts and images.
Unlike the above-mentioned studies, \citeA{DBLP:conf/aaai/HinamiIYM21} apply TIMT to the comics domain. They extract semantic tags of each comic scene as image context and prepend these tags to the source text.

In complex and realistic scenarios, cascaded translation systems exhibit superior performance. Different from the above-mentioned end-to-end approaches, \citeA{DBLP:conf/acl/LanYLZ0WHS23} present a cascaded TIMT model with a multi-modal codebook, which can leverage the input image to generate latent codes encoding the information of relevant or correct texts, thus providing useful supplementary information to alleviate the OCR error propagation. Moreover, they propose a multi-stage training framework that makes full use of additional bilingual texts and OCR data to enhance the model training.

\subsection{Video-guided Machine Translation}
In this task, given pairs of source text and video, the translation models are required to automatically generate the target translations by combining both textual and video information. This task has many real-world applications, such as subtitle translation for social media.
Similar to the aforementioned types of MMT tasks, the studies on this task also face many challenges. First, this task suffers from data scarcity. Second, compared with images, videos provide richer visual information such as actions and temporal transitions, while also introducing some visual redundancy. Therefore, how to effectively extract and utilize video information is an important challenge in this task.

The studies on this task can be roughly divided into two categories: model design and dataset construction. \citeA{DBLP:conf/iccv/WangWCLWW19} first explore the task of video-guided MT, where they propose a multi-modal sequence-to-sequence model with two attention mechanisms to respectively capture textual and video information. Considering the order information in video frames, \citeA{DBLP:journals/corr/abs-2006-12799} add positional embeddings to keyframe-based video representations. Besides, they employ not only a video encoder to capture motion features in videos, but also an image encoder to capture object and scene features in each keyframe. Note that previous studies mainly focus on the representations of video motions to solve the verb sense ambiguity in the source text, leaving the noun sense ambiguity unsolved. To deal with this issue, \citeA{DBLP:conf/acl/GuSCK21} propose a spatial hierarchical attention module that utilizes the spatial representations in input videos. Concretely, they first apply an object-level attention layer to summarize the object-level spatial representations into frame-level spatial representations, and then employ another frame-level attention layer to summarize all ordered frame-level spatial representations into video-level representations. 
Along this line, \citeA{DBLP:conf/emnlp/LiSCKL23} introduce two training objectives based on the selective attention model \shortcite{DBLP:conf/acl/LiLZZXMZ22}: 1) the frame attention loss allowing the model to focus more on the central frames where the subtitles occur, 2) the ambiguity augmentation loss that enables the model to pay more attention to the possibly-ambiguous data.
\citeA{DBLP:conf/acl/KangHP0SCWHS23} introduce a video-guided MT model with a cross-modal encoder. This model is trained with a contrastive learning objective, which brings close the representations of the matched source texts and videos, while pushing the representations of the unrelated source texts and videos farther.
To better exploit video information, \citeA{DBLP:journals/pami/LiHCHYH23} employ a spatial-temporal graph network \shortcite{DBLP:conf/cvpr/PanCHLGAN20} to capture object information among frames in videos. They define four tasks for unsupervised training: 1) unsupervised video-guided MT, which pulls close the input text and the output of the back-translation model with the video as a pivot, 2) video-text matching, which can map textual and video representations into a modality-shared semantic space, 3) video captioning for paired-translation, which feeds the video into pre-trained video captioning models of different languages to generate source-target pseudo text pairs for translation, 4) video captioning for back-translation, which pulls close the texts generated by video captioning models and the outputs of the back-translation models.

To overcome data scarcity, some studies \shortcite{DBLP:journals/corr/abs-1811-00347,DBLP:conf/iccv/WangWCLWW19,DBLP:conf/acl/KangHP0SCWHS23} release several multi-modal video datasets. The How2 dataset \shortcite{DBLP:journals/corr/abs-1811-00347}  consists of videos, utterance-level English subtitles, aligned Portuguese translations, and video-level English summaries. To increase the diversity of video captions, each video in the VaTeX dataset \shortcite{DBLP:conf/iccv/WangWCLWW19} is equipped with 10 English and 10 Chinese descriptions. \citeA{DBLP:conf/acl/KangHP0SCWHS23} release the BigVideo dataset, whose size is an order of magnitude larger than the size of the previous largest available dataset, and \citeA{DBLP:conf/emnlp/LiSCKL23} release the large-scale EVA dataset in the movie and TV domain. 

\subsection{Multi-modal Simultaneous Machine Translation}
Similar to traditional simultaneous machine translation (SiMT), the speech is first converted to the source texts in this task, and then the translation models are required to generate the target translations based on the continuous source text streams and given images. This task can be applied to some realistic applications where images are presented before the complete source text streams are available, such as presentations with slides and news video broadcasts. 
The primary challenge of this task lies in how to effectively use image information as the supplementary context to enrich incomplete textual information, so as to obtain target translations with low latency and high quality.

In this respect, \shortcite{DBLP:conf/wmt/ImankulovaKHK20} is the first attempt to explore the multi-modal SiMT task, where a hierarchical attention mechanism \shortcite{DBLP:conf/acl/LibovickyH17} is used to incorporate textual and image information. Following the previous works \shortcite{DBLP:conf/wmt/CalixtoEF16,DBLP:journals/corr/abs-1712-03449}, \citeA{DBLP:conf/emnlp/CaglayanIHMBS20} integrate image information into the encoder or decoder modules through a multi-modal attention mechanism. Experimental results show that utilizing image information can provide the model with the missing source context, allowing it to correctly translate the gender-marked words and deal with the differences in word order. Besides, regional image features are more effective than global image features in this task. \citeA{DBLP:conf/eacl/IveLMCMS21} further introduce reinforcement learning to this task. They propose three strategies to integrate image information: 1) using image features to initialize the agent network, 2) applying a multi-modal attention mechanism to generate the image context vector in the agent network, 3) taking a MMT model as the environment network. In contrast to previous methods that use RNN networks, \citeA{DBLP:journals/jair/HaralampievaCS22} pioneer the use of Transformer architectures. Apart from the main multi-modal SiMT task, they design an auxiliary training task where the visual attention mechanism is supervised by annotated phrase-region alignments, so that the additional image information can better complement the missing source context.

\subsection{Multi-modal Chat Machine Translation}
Conversations in real-life scenarios often involve multi-modal information and their content largely depends on the scenes that speakers observe.
Therefore, visual information can be a good supplement to dialogue history context. In this task, the model leverages both bilingual dialogue history contexts and the associated visual context to translate the current source utterance. This task finds its applications in subtitle translation for movies and TV episodes, especially some conversational scenes.
In real life, conversations usually involve multi-sense words and pronominal anaphora issues. Thus, how to efficiently utilize visual information to resolve these problems remains a challenge.

\shortcite{DBLP:conf/acl/LiangMXCZ22} is the first attempt to explore this task. In this work, the authors create a Multimodal Sentiment Chat Translation Dataset (MSCTD), aiming to generate more accurate dialogue translations with the guidance of additional visual context. This dataset includes 17,841 multi-modal bilingual conversations, each consisting of multiple quadruples in the format of \textlangle \emph{English utterance, Chinese/German utterance, image, sentiment}\textrangle. Based on this dataset, they adapt existing MMT models and textual chat translation models to construct several benchmarks for this task. Experimental results demonstrate that integrating image information indeed improves the quality of dialogue translations.

\begin{table}[ht]
\newcommand{\tabincell}[2]{\begin{tabular}{@{}#1@{}}#2\end{tabular}}
\tiny
\centering
\begin{threeparttable}[width=0.3\textwidth]
\setlength{\tabcolsep}{0.8mm}{
    \renewcommand{\arraystretch}{1.5}
    \begin{tabular}{l|c|c|c|c|c|c|c|c@{}}
    \toprule
    \textbf{Dataset} & \textbf{Sub-Dataset} & \textbf{Ambigious} & \textbf{Domain} & \textbf{Task} & \textbf{Lauguage} & \textbf{Image} & \textbf{Video} & \textbf{Text}  \\ 
    \midrule
    
    IAPR TC-12 & - & \ding{55} & Daily Activity & S-MT & \multicolumn{1}{c|}{DE, EN} & 20.0K & - & 20.0K \\ \hline
    Multi30K & - & \ding{55} & Daily Activity & S-MT, Si-MT & \multicolumn{1}{c|}{CS, DE, EN, FR} & 31.0K & - & 31.0K \\  \hline
    
    
    \multirow{2}{*}{MLT} & \multirow{2}{*}{-} & \multirow{2}{*}{\ding{51}} & \multirow{2}{*}{Daily Activity} & \multirow{2}{*}{S-MT} &  \multicolumn{1}{c|}{DE, EN} & 53.9K & - & 53.9K \\ 
    & & & & & \multicolumn{1}{c|}{EN, FR} & 44.8K & - & 44.8K \\ \hline
    
    
    MultiSense & - & \ding{51} & Daily Activity & S-MT & \multicolumn{1}{c|}{DE, EN, ES} & 9.5K & - & 9.5K \\  \hline
    
    AmbigCaps & - & \ding{51} & Daily Activity & S-MT & \multicolumn{1}{c|}{EN, TR} & 91.6K & - & 91.6K \\  \hline
    
    \multirow{2}{*}{$\mathrm{M}^{3}$} & ${\rm {M}^{3}}$-Multi30K & \ding{55} & \multirow{2}{*}{Daily Activity} & \multirow{2}{*}{S-MT} & \multirow{2}{*}{CS, DE, EN, FR, HI, LV, TR} & 31.0K & - & 31.0K \\ 
    & ${\rm {M}^{3}}$-AmbigCaps & \ding{51} &  &  &  & 91.6K & - & 91.6K \\  \hline
    
    \multirow{2}{*}{Fashion-MMT} & Fashion-MMT(L) & \ding{55} & \multirow{2}{*}{E-commerce} & \multirow{2}{*}{E-MT} & \multicolumn{1}{c|}{EN, ZH} & 885.2K & - & 114.3K \\
    & Fashion-MMT(C) & \ding{55} &  &  & \multicolumn{1}{c|}{EN, ZH} & 312.7K & - & 40.0K \\  \hline
    
    EMMT & - & \ding{55} & E-commerce & E-MT & \multicolumn{1}{c|}{EN, ZH} & 22.0K & - & 875.0K  \\  \hline
    
    \multirow{3}{*}{TIT Dataset} & \multirow{3}{*}{-} & \multirow{3}{*}{\ding{55}} & \multirow{3}{*}{-} & \multirow{3}{*}{TIMT} & \multicolumn{1}{c|}{ZH$\Rightarrow$EN} & 1.0M & - & 1.0M \\ & & & & & \multicolumn{1}{c|}{EN$\Rightarrow$ZH} &  1.0M & - & 1.0M  \\  & & & & & \multicolumn{1}{c|}{EN$\Rightarrow$DE} &  1.0M & - & 1.0M  \\ \hline
    
    BLATID & - & \ding{55} & - & TIMT & \multicolumn{1}{c|}{EN, ZH} & 1.2M & - & 1.2M \\  \hline
    
    OCRMT30K & - & \ding{55} & - & TIMT & \multicolumn{1}{c|}{EN, ZH} & 30.2K & - & 164.7K \\  \hline
    
    How2 & - & \ding{55} & Social Media & V-MT & \multicolumn{1}{c|}{EN, PT} & - & 191.6K & 191.6K \\  \hline
    
    VaTeX & - & \ding{55} & Social Media & V-MT & \multicolumn{1}{c|}{EN, ZH} & - & 41.3K & 412.7K \\  \hline
    
    BigVideo & - & \ding{55} & Social Media & V-MT & \multicolumn{1}{c|}{EN, ZH} & - & 4.5M & 4.5M \\  \hline
    
    \multirow{2}{*}{VISA} & {VISA-Polysemy} & \multirow{2}{*}{\ding{51}} & \multirow{2}{*}{Movie and TV} & \multirow{2}{*}{V-MT} & \multirow{2}{*}{EN, JA} & - & 20.7K & 20.7K \\ & {VISA-Omission} &  &  & &  & - & 19.2K & 19.2K \\ \hline
    
    \multirow{2}{*}{EVA} & \multirow{2}{*}{-} & \multirow{2}{*}{\ding{51}} & \multirow{2}{*}{Movie and TV} & \multirow{2}{*}{V-MT} & \multicolumn{1}{c|}{EN, JA} & - & 852.4K & 852.4K \\ & & & & & \multicolumn{1}{c|}{EN, ZH} & - & 519.7K & 519.7K \\  \hline
    
    \multirow{2}{*}{MSCTD} & \multirow{2}{*}{-} & \multirow{2}{*}{\ding{55}} & \multirow{2}{*}{Dialogue} & \multirow{2}{*}{C-MT} & \multicolumn{1}{c|}{EN, ZH} & 142.9K & - & 142.9K \\ & & & & & \multicolumn{1}{c|}{DE, EN} & 30.4K & - & 30.4K \\  \hline
    
    BIG-C & - & \ding{55} & Dialogue & C-MT & \multicolumn{1}{c|}{BE, EN} & 16.2K & - & 92.1K \\  \hline
    
    HaVQA & - & \ding{55} & QA & Q-MT & \multicolumn{1}{c|}{EN, HA} & 1.6K & - & 12.0K  \\  
    
    \bottomrule
    \end{tabular}
 }
 \end{threeparttable}
 \caption{Summary statistics from commonly-used MMT datasets. Note that Ambiguous column refers to whether the dataset contains ambiguous words, and video refers to video clips. S-MT, Si-MT, V-MT, C-MT, Q-MT, E-MT refer to scene-image MMT, multi-modal SiMT,  video-guided MT, multi-modal chat MT, multi-modal QA MT, e-commerce product-oriented MT, respectively. Twelve languages are covered in these datasets: English (EN), French (FR), German (DE), Spanish (ES), Czech (CS), Turkish (TR), Hindi (HI), Latvian (LV), Japanese (JA), Portuguese (PT), Chinese (ZH), Bemba (BE), Hausa (HA).}
\label{tab:Datasets}
 \end{table}

\section{Datasets}
Table~\ref{tab:Datasets} shows the information of the commonly used datasets used in MMT. All datasets are English-centric translation corpora. They cover a diverse range of domains, including daily activity, movie and TV, social media, e-commerce and QA. In the following, we will introduce these datasets in terms of data source, data quantity, data composition and so on.

\emph{IAPR TC-12} \shortcite{Grubinger2006TheIT}. This dataset consists of 20,000 images from a private photographic image collection, which involves many categories such as sports, actions, photographs of people, animals, cities, landscapes and many other aspects of contemporary life. In this dataset, each image is annotated with a German description and its English translation. Besides, the dataset contains some additional annotations such as titles and locations in German, English and Spanish.

\emph{Multi30K} \shortcite{DBLP:conf/acl/ElliottFSS16}. As the most frequently used dataset in the scene-image MMT task, Multi30K is extended from Flickr30K \shortcite{DBLP:journals/tacl/YoungLHH14}, where images are about human daily activities and come from online photo-sharing websites. It contains 31,014 images and each image is paired with an original English description and its German-French-Czech translations. Apart from the translations, it also contains non-parallel English and German descriptions for each image.

\emph{MLT} \shortcite{DBLP:conf/lrec/LalaS18}. This dataset is generated from Multi30K, consisting of 53,868 samples for English to German and 44,779 samples for English to French. Each sample is a quadruple in the format of \textlangle \emph{source ambiguous word, target word, source sentence, image}\textrangle. 

\emph{How2} \shortcite{DBLP:journals/corr/abs-1811-00347}. Unlike other single-task datasets, How2 crawls videos along with various types of metadata from YouTube, obtaining 2,000 hours of videos in total. Among the crawled videos, 300 hours of them are paired with utterance-level English subtitles, aligned Portuguese translations and video-level English summaries, covering 22 topics. Therefore, this dataset can be widely used for various multi-modal tasks, including automatic speech recognition, speech-to-text translation and MMT.

\emph{VaTeX} \shortcite{DBLP:conf/iccv/WangWCLWW19}. This dataset is a large-scale English-Chinese video captioning dataset, where videos come from the widely-used action classification dataset, Kinetics-600 \shortcite{DBLP:journals/corr/KayCSZHVVGBNSZ17}. It contains 41,269 video clips and  825,380 captions in total, and these video clips of the train and validation sets are labeled with 600 fine-grained action labels. Moreover, to increase the caption diversity, each video clip in VaTeX is annotated with 10 English and 10 Chinese descriptions, half of which are independent annotations and the other half are paired translations of each other.

\emph{MultiSense} \shortcite{DBLP:conf/naacl/GellaEK19}. This dataset contains 9,504 images annotated with ambiguous English verbs and their context-consistent translations in German and Spanish.  Additionally, the authors annotate a subset of 995 \textlangle \emph{English description, German translation, image}\textrangle \enspace triplets.

\emph{AmbigCaps} \shortcite{DBLP:conf/emnlp/LiAS21}. It is a gender-ambiguous dataset containing 91,601 sentences automatically translated from English to Turkish via Google Translate and their associated images. Particularly, it filters out the sentences in Conceptual Captions \shortcite{DBLP:conf/acl/SoricutDSG18} that contain nouns with gender information or professions referring to one gender.

\emph{Fashion-MMT} \shortcite{DBLP:conf/mm/0003CJLXH21}. It is the first public bilingual product description dataset, which is based on the fashion captioning dataset FACAD \shortcite{DBLP:conf/eccv/YangZJLWTXWW20}. In this dataset, each description is aligned with an average of 6 to 7 product images of different colors and poses. Product categories and attribute labels are also provided for each product.
Most importantly, there are two types of translations provided, forming two sub-datasets. The first one is a noisy version, denoted as Fashion-MMT(L), which contains 114,257 automatic Chinese translations of original English product descriptions via Google Translate. The second one is a clean version, denoted as Fashion-MMT(C), containing 40,000 samples with manually annotated Chinese translations.

\emph{MSCTD} \shortcite{DBLP:conf/acl/LiangMXCZ22}. This dataset is proposed for multi-modal chat translation. To build MSCTD, researchers select the multi-modal dialogs from the OpenViDial dataset \shortcite{DBLP:journals/corr/abs-2109-12761}. In total, it contains 17,841 bilingual conversations, where each original English utterance is paired with a Chinese/German translation, an image depicting the current conversational scene and a sentiment label.

\emph{${ M^{3} }$} \shortcite{DBLP:conf/emnlp/GuoLHYL0C22}. There are two multi-lingual versions in ${ \rm M^{3} }$: ${ \rm M^{3} }$-Multi30K and ${ \rm M^{3}}$-AmbigCaps. ${ \rm M^{3} }$-Multi30K is an extension of the existing Multi30K dataset with additional Turkish, Hindi, Latvian translations, and ${ \rm M^{3} }$-AmbigCaps is an extension of the existing AmbigCaps dataset with additional French, Czech, Turkish, Hindi, Latvian translations.

\emph{VISA} \shortcite{DBLP:conf/lrec/LiSGCK22}. This dataset consists of 39,880 Japanese-English bilingual subtitles and corresponding video clips from movies and TV episodes. Particularly, the Japanese subtitles are ambiguous, and the whole dataset is divided into Polysemy and Omission according to the causes of ambiguity.

\emph{Text Image Translation (TIT) Dataset} \shortcite{DBLP:conf/icpr/MaZTHWZ022}. This dataset comprises a synthetic text image dataset for training and two real-world datasets for evaluation, including subtitle and street-view test sets. The synthetic text image dataset considers three language pairs: English$\Rightarrow$Chinese, English$\Rightarrow$German, and Chinese$\Rightarrow$English, with each pair containing 1 million training samples and 2,000 validation samples and each sample consisting of a source image and a target text. Additionally, the subtitle test set contains 1,040 samples, while the street-view test set has 1,198 samples.

\emph{BLATID} \shortcite{DBLP:journals/tmm/ChenYYL23}. BLATID is a Chinese-English bilingual annotation TIMT dataset generated from the existing MT corpus, AIC\footnote{\url{https://github.com/AIChallenger/AI_Challenger_2018}}. It contains 1 million training samples, 100K validation samples and 55K test samples. Especially, the test samples are derived from movies and their related bilingual subtitles. Each sample in this dataset is made up of a Chinese source image, a Chinese source text along with an English target text.

\emph{OCRMT30K} \shortcite{DBLP:conf/acl/LanYLZ0WHS23}. Previous studies on the TITM task mainly center around constructing synthetic TITM datasets, which are far from the real scenarios. To address this issue, OCRMT30K is annotated over five commonly used Chinese OCR datasets \shortcite{DBLP:conf/icdar/ShiYLYXCBLB17,DBLP:journals/tip/HeZYL18,DBLP:conf/icdar/NayefLOPBCKKM0B19,DBLP:conf/iccv/SunLLHDL19,DBLP:conf/icdar/ChngDLKCJLSNLNF19} where the images are freely captured in the streets. It totally includes 30,186 images and 164,674 Chinese-English parallel texts.

\emph{EMMT} \shortcite{DBLP:conf/acl/ZhuSCHWW23}. This dataset is an English-Chinese e-commerce MMT dataset crawled from TikTok Shop and Shoppee. It incorporates three types of data, including 22K bilingual texts with images, 750K parallel texts and 103K monolingual captions. Particularly, 500 samples that contain ambiguous words and will be translated mistakenly without considering the image information are carefully selected as the test set.

\emph{BigVideo} \shortcite{DBLP:conf/acl/KangHP0SCWHS23}. Consisting of 4.5 million sentence pairs and 9,981 hours of videos, BigVideo is the largest English-Chinese video subtitle dataset to date, where videos are collected from two popular online video platforms, YouTube and Xigua. Each video clip in this dataset is paired with a source subtitle and its translation. Besides, two test sets are introduced to verify the necessity of visual information: AMBIGUOUS and UNAMBIGUOUS. The former contains 877 samples with the presence of ambiguous words, while the latter contains 1,517 samples where the textual context is sufficient for translation.

\emph{EVA} \shortcite{DBLP:conf/emnlp/LiSCKL23}. This dataset is the largest video subtitle translation dataset in the movie and TV domain, containing 852,440 Japanese-English parallel subtitle pairs, 519,673 Chinese-English parallel subtitle pairs, and corresponding video clips collected from movies and TV episodes. In particular, the source subtitles in its evaluation set are ambiguous and the corresponding videos are guaranteed to be helpful for disambiguation.

\emph{BIG-C} \shortcite{DBLP:conf/acl/SikasoteMAA23}. This dataset is a multi-modal one in Bemba, which can be applied to many NLP tasks. It is made up of multi-turn dialogues between Bemba speakers, totally containing 16,229 images, 92,117 utterances and 185 hours of audio data. Each sample consists of an image, audio data grounded on the image, its dialogue's corresponding audio transcriptions and English translations. 

\emph{HaVQA} \shortcite{DBLP:conf/acl/ParidaAMBKAKSBK23}. It is the first multi-modal dataset for various multi-modal tasks in Hausa, such as MMT, VQA and visual question elicitation. Totally, this dataset provides 1,555 images and 6,020 questions/answers in both English and Hausa.

In summary, compared with existing large-scale bilingual corpora, these datasets suffer from relatively smaller sizes, limited language pairs and domains. These factors seriously constrain the applications of existing MMT models. Therefore, we expect the emergence of large-quantity and high-quality datasets in the future, which will significantly prompt the development of MMT.

\section{Evaluation}
The commonly used automatic evaluation metrics for  MMT include BLEU \shortcite{DBLP:conf/acl/PapineniRWZ02}, METEOR \shortcite{DBLP:conf/wmt/LavieA07,DBLP:conf/wmt/DenkowskiL14}, TER \shortcite{DBLP:conf/amta/SnoverDSMM06}, chrF \shortcite{DBLP:conf/wmt/Popovic15}, CIDEr \shortcite{DBLP:conf/cvpr/VedantamZP15}, and COMET \shortcite{DBLP:conf/emnlp/ReiSFL20}, all of which have been widely used in text-only MT. In the following, we will provide a detailed description of each metric.
\begin{itemize}
\item \emph{BLEU}. As the most commonly used evaluation metric in MT, BLEU aims to measure the precision of n-gram matches between translation and reference at the word level. It possesses the advantages of efficient and convenient computation as well as considering n-gram information. However, it also has limitations such as disregarding the grammatical correctness of translations, exhibiting a bias towards shorter translations, and not effectively dealing with translations involving synonyms or conveying the same meaning.
\item \emph{TER}. It is a distance-based evaluation metric, which assesses the quality of a translation by calculating its minimum number of edits to the reference.
\item \emph{chrF}. Unlike BLEU, chrF measures the overlap of n-grams between translation and reference at the character level. Besides, it considers the morphological complexity of languages (e.g., different tenses).
\item \emph{METEOR}. To address some limitations of BLEU, METEOR comes up with a relaxed matching strategy to perform exact word, stem word, synonym and paraphrase matching, and considers precision more than recall. 
\item \emph{CIDEr}. This metric is specifically designed for image captioning, where each image is paired with multiple reference captions. It uses Term-Frequency Inverse Document Frequency (TF-IDF) \shortcite{DBLP:journals/jd/Robertson04} to weigh each n-gram in the translation and reference.
Unlike previous metrics, CIDEr distinguishes the importance of different n-grams through TF-IDF weights and focuses more on whether keywords in the reference are present in the translation.
\item \emph{COMET}. In comparison to the above string-based metrics (BLEU, TER, chrF, METEOR), COMET is a neural network based metric that supports multiple languages, achieving better evaluation results than conventional metrics. Concretely, it utilizes XLM-RoBERTa \shortcite{DBLP:conf/nips/ConneauL19} for model initialization and takes the source text, translation, and reference as inputs to compute a score, which accounts for the semantic similarity among these inputs.
\end{itemize}

Notice that the above-mentioned metrics perform evaluation by assessing the similarity between translation and reference, and thus the quality of references becomes a crucial factor impacting the reliability of evaluation results. We expect more evaluation metrics that are not limited by references and effectively integrate the unique characteristics of MMT in the future.

In addition to academic papers published in conferences and journals, shared tasks play a significant role in promoting the development of  MMT. One prominent event in this regard is the annual Workshop on Machine Translation (WMT)\footnote{\url{https://machinetranslate.org/wmt}}, which involves various tasks related to MT, such as MMT tasks\footnote{\url{https://www.statmt.org/wmt16/multimodal-task.html}}. From 2016 to 2018, WMT organizes three shared tasks of MMT and summarizes submissions from participants around the world \shortcite{DBLP:conf/wmt/SpeciaFSE16,DBLP:conf/wmt/ElliottFBBS17,DBLP:conf/wmt/BarraultBSLEF18}. \citeA{DBLP:conf/wmt/SpeciaFSE16} explore the MMT task for the first time, where the Multi30K dataset is used and BLEU, METEOR, TER are employed as evaluation metrics. They conclude that the neural networks based MMT models do not perform as well as the text-only SMT models in any of the submissions. Following \citeA{DBLP:conf/wmt/SpeciaFSE16}, \citeA{DBLP:conf/wmt/ElliottFBBS17} add additional French translations to Multi30K and construct two new evaluation sets: Test2017 and Ambiguous COCO (MSCOCO) which contains ambiguous verbs in the source language. Their work highlights some improvements in the performance of MMT models compared to last year, emphasizing that the incorporation of external resources can further enhance these models. Moreover, \citeA{DBLP:conf/wmt/BarraultBSLEF18} extend Multi30K to include another new language, Czech. They also propose a novel evaluation metric called Lexical Translation Accuracy (LTA), which measures the accuracy of translating ambiguous words. In 2018, almost all submitted models achieved better results compared to the text-only SMT. 
In conclusion, the MMT shared tasks have played a pivotal role in enriching the associated datasets, refining evaluation metrics, and fostering gradual improvements in the performance of the models proposed by diverse teams.

Prior studies show that image context is only needed in some specific scenarios, such as translating incorrect or ambiguous words or gender-neutral words that need to be marked for gender in the target language \shortcite{DBLP:journals/nle/FrankES18}.   
In order to investigate the impact of image context in different scenarios, some researchers create samples with limited textual context by designing various input degradation strategies to mask words with specific attributes in the source text, such as random words, ambiguous words, gender-neutral words, color words, entity words, words referring to people (e.g.,\emph{``man''} and \emph{``woman''})  \shortcite{DBLP:conf/acl/IveMS19,DBLP:conf/naacl/CaglayanMSB19,DBLP:conf/acl/WuKBLK20,DBLP:conf/aaai/WangX21,DBLP:conf/acl/FangF22,DBLP:conf/cvpr/LiPKCFCV22,DBLP:conf/emnlp/PengZZ22,DBLP:conf/acl/LiLZZXMZ22}. 
Other researchers also explore whether image context can help eliminate gender ambiguity by evaluating gender accuracy in the translations of gender-neutral languages \shortcite{DBLP:conf/emnlp/LiAS21}.

\section{Comparison between Existing Models}
In this section, we summarize the experimental results of some representative MMT models. Early research predominantly employs RNN architectures, while more and more recent studies have shifted towards Transformer architectures. Since there are a lot of studies focusing on scene-image MMT and their experimental results indicate that the performance of Transformer-based models is usually better than that of RNN architectures, we only present experimental results of Transformer-based models in the scene-image MMT task. Notice that experimental configurations of different studies are not exactly the same, such as image encoders, evaluation datasets and metrics, so direct and quantitative comparison between these studies is challenging. We can only infer the relative advantages of each model and identify the factors that influence translation performance.

The experimental results of the scene-image MMT task are shown in Table~\ref{tab:SMT},  from which we can draw the following conclusions:
\begin{itemize}
\item Early works tend to use ResNet \shortcite{DBLP:conf/cvpr/HeZRS16} and Faster-RCNN \shortcite{DBLP:conf/nips/RenHGS15} as image encoders to extract global and regional image features. However, in recent years, an increasing number of studies opt for more complicated image encoders, such as CLIP \shortcite{DBLP:conf/icml/RadfordKHRGASAM21} or Vision Transformer \shortcite{DBLP:conf/iclr/DosovitskiyB0WZ21,DBLP:conf/iccv/LiuL00W0LG21}, to capture richer image semantic information.

\item Incorporating various performance-boosting techniques can result in high-quality translations. Notably, Transformer-based models that adopt contrastive learning \shortcite{DBLP:conf/cikm/ChengZLLZ23,DBLP:conf/ijcnn/WangZGYL22}, double attention mechanism \shortcite{DBLP:journals/ijon/ZhaoKKC22,DBLP:conf/acl/IveMS19}, multi-task learning \shortcite{DBLP:conf/emnlp/ZuoLLZXZ23,DBLP:conf/acl/0001LZZC23} and pre-training \shortcite{DBLP:conf/iccv/GuptaKZ0P023,DBLP:conf/acl/KongF21} techniques exhibit better performance.

\item In most real-world scenarios, images paired with source texts are not always available during inference. Therefore, more and more studies concentrate on image-free scenarios, leveraging imagination, image retrieval and unsupervised learning techniques to achieve high-quality translations.
\end{itemize}

In summary, for the scene-image MMT task, the utilization of effective image encoders and diverse performance-boosting techniques can lead to high-performance models. Table~\ref{tab:VMT} and Table~\ref{tab:SiMT} show experimental results of the video-guided MT and the multi-modal SiMT tasks, respectively.\footnote{Since there are few studies on the e-commerce product-oriented MT and the multi-modal chat MT tasks, we do not present their experimental results here. Besides, due to data scarcity, the studies on the TIMT task usually conduct experiments on the self-constructed datasets, which prevents us from comparing these works. Therefore, we also omit the experimental results of the TIMT task. } 
Generally, the models integrating visual information yield superior performance compared to the text-only models.


\begin{sidewaystable}[thp]
\scriptsize
\centering
\begin{threeparttable}[width=0.5\textwidth]
\setlength{\tabcolsep}{1.2mm}{
 \begin{tabular}{p{3.0cm}|c|c|c|ccc|ccc}
    \toprule
    \multirow{3}{*}{\textbf{Model}} & \multirow{3}{*}{\textbf{Image Encoder}} & \multirow{3}{*}{\textbf{Technique}} & \multirow{3}{*}{\textbf{Image-free}} & \multicolumn{3}{c|}{\textbf{EN$\Rightarrow$DE}} & \multicolumn{3}{c}{\textbf{EN$\Rightarrow$FR}} \\ [3pt] \cline{5-10} \rule{0pt}{3ex}
    & & &  & \textbf{Test16} & \textbf{Test17}& \textbf{MSCOCO}& \textbf{Test16}& \textbf{Test17}& \textbf{MSCOCO}\\
    & & &  & \textbf{B/M}& \textbf{B/M}& \textbf{B/M}& \textbf{B/M}& \textbf{B/M}& \textbf{B/M}\\ \midrule
    \shortcite{DBLP:conf/acl/IveMS19} & ResNet & DA+MD  & \ding{55} & 38.0/55.6& - & - & 60.1/74.6  & - & - \\
    \shortcite{DBLP:conf/cvpr/SuFBKH19} & ResNet & ML+PT+UL  & \ding{51} & 23.5/26.1& - & - & 39.8/35.5  & - & - \\
    \shortcite{DBLP:conf/acl/YinMSZYZL20} & Faster R-CNN  & GNN& \ding{55} & 39.8/57.6& 32.2/51.9  & 28.7/47.6  & 60.9/74.9  & 53.9/69.3  & - \\
    \shortcite{DBLP:conf/acl/YaoW20} & ResNet & BT+CFM & \ding{55} & 39.5/56.9& - & - & - & - & - \\
    \shortcite{DBLP:conf/mm/LinMSYYGZL20} & Fast R-CNN+ResNet & CN & \ding{55} & 39.7/56.8& 31.0/49.9  & 26.7/45.7  & 61.2/76.4  & 54.3/70.3  & 45.4/65.0  \\
    \shortcite{DBLP:conf/iclr/0001C0USLZ20} & ResNet & GF+IR  & \ding{51} & 35.7/-& 26.9/-  & - & 58.3/-  & 48.7/-  & - \\
    \shortcite{DBLP:conf/aaai/YangCZ020} & Faster R-CNN  & ML & \ding{55} & -  & 29.5/50.3  & - & - & 53.3/70.4  & - \\
    \shortcite{DBLP:conf/coling/NishiharaTNON20} & ResNet & ML & \ding{55} & 40.5/59.1& - & - & - & - & - \\
    \shortcite{DBLP:conf/acl/HuangHCH20} & Fast R-CNN & ML+PT+UL  & \ding{51} & 33.9/54.1& - & - & 52.3/67.6  & - & - \\
   \shortcite{DBLP:conf/naacl/LongWL21}  & GAN & AT+IG  & \ding{51} & 38.6/55.7& 32.4/52.5  & 28.8/48.9  & 59.9/74.3  & 52.8/68.6  & 45.3/65.1  \\
    \shortcite{DBLP:conf/emnlp/HuangZZ21}  & ResNet & ML+MU  & \ding{55} & 39.7/57.8& 32.9/52.1  & 29.1/47.5  & - & - & - \\
   \shortcite{DBLP:conf/eacl/CaglayanKAMEES21}  & Faster R-CNN  & PT & \ding{55} & 44.0/61.3& 38.1/57.2  & 35.2/53.8  & - & - & - \\
   \shortcite{DBLP:conf/acl/KongF21}  & Faster R-CNN  & PT & \ding{55} & 42.7/60.7& 35.5/54.9  & 32.8/52.2  & 65.8/79.1  & 58.2/73.5  & - \\
   \shortcite{DBLP:conf/acl/WuKBLK20} & ResNet & GF & \ding{55} & 42.0/-& 33.6/-  & 29.0/-  & 61.7/-  & 54.9/-  & 44.9/-  \\
   \shortcite{DBLP:journals/ijon/ZhaoKKC22}  & Fast R-CNN+ResNet & DA & \ding{55} & 38.6/57.7& - & - & 60.1/75.0  & - & - \\
    \shortcite{DBLP:journals/taslp/ZhaoKKC22} & Fast R-CNN+ResNet & CFM+DA & \ding{55} & 39.3/58.3& 32.3/52.8  & 28.5/48.5  & 61.8/76.3  & 54.1/70.6  & 43.4/63.8  \\
   \shortcite{DBLP:conf/coling/YeGXT022} & ResNet & CFM+GF & \ding{55} & 42.6/60.0& 35.1/54.5  & 31.1/50.5  & 63.2/77.5  & 55.5/72.6  & 46.3/67.4  \\
   \shortcite{DBLP:journals/apin/YeG22}  & \multicolumn{1}{l|}{Fast R-CNN+ResNet}  & GF+MU  & \ding{55} & 41.8/58.9& 33.1/51.9  & 29.9/49.1  & 62.2/76.9  & 55.2/73.4  & 44.4/66.4  \\
   \shortcite{DBLP:conf/cvpr/LiPKCFCV22}  & VQGAN-VAE  & IG & \ding{51} & 42.6/69.3& 35.1/62.8  & 30.7/57.6  & 63.1/81.8  & 56.0/77.1  & 46.4/71.3  \\
   \shortcite{DBLP:conf/emnlp/PengZZ22}  & ResNet & IG+KD  & \ding{51} & 41.3/58.9& 33.8/53.2  & 30.2/48.9  & 62.5/77.2  & 54.8/71.9  & - \\
   \shortcite{DBLP:conf/acl/FangF22}  & Faster R-CNN  & GF+IR  & \ding{51} & 40.3/-& 33.5/-  & 30.3/-  & 61.3/-  & 53.2/-  & 43.7/-  \\
   \shortcite{DBLP:conf/mir/PengZZ22} & ResNet & GF+ML  & \ding{55} & 42.2/59.4& 34.1/53.1  & 30.8/49.3  & 63/77.3 & 55.5/72.4  & - \\
   \shortcite{DBLP:conf/ijcnn/WangZGYL22}  & ResNet & CT & \ding{55} & 42.2/68.8& 34.3/62.1  & 30.6/57.3  & 62.8/81.4  & 54.7/76.4  & 45.5/71.0  \\
   \shortcite{DBLP:conf/emnlp/JiZZHS22} & ResNet & CT+GF  & \ding{55} & 41.8/68.6& 34.6/62.4  & 30.6/56.7  & - & - & - \\
   \shortcite{DBLP:conf/acl/LiLZZXMZ22}  & Vision Transformer & CFM+GF & \ding{55} & 41.8/68.6& 34.3/62.3  & 30.2/56.9  & 62.2/81.4  & 54.5/76.3  & 44.8/70.6  \\
   \shortcite{DBLP:conf/acl/YuasaTKNK23}  & Vision Transformer & CFM+GF+SI & \ding{55} & 41.2/-& 32.2/-  & 28.3/-  & - & - & - \\
    \shortcite{DBLP:conf/acl/FuteralSLSB23} & MDETR+CLIP & ML+PT  & \ding{55} & 43.3/-& 38.3/-  & 35.7/-  & 67.2/-  & 61.6/-  & 51.1/-  \\
     \shortcite{DBLP:conf/acl/ZhuSCHWW23} & X-VLM& GF+ML  & \ding{55} & 41.0/-& 34.6/-  & - & - & - & - \\
     \shortcite{yin2023multi} & Faster R-CNN  & CT+GNN & \ding{55} & 40.5/58.4& 33.2/52.3  & 29.6/48.8  & 61.6/76.0  & 54.4/70.7  & - \\
    \shortcite{DBLP:conf/acl/0001LZZC23} & Faster R-CNN  & ML+UL  & \ding{51} & 37.4/57.2& - & - & 56.9/70.7  & - & - \\
    \shortcite{DBLP:conf/iccv/GuptaKZ0P023} & M-CLIP & PT & \ding{51} & 43.9/70.2& 37.2/65.4  & 34.5/61.3  & 64.6/82.5  & 57.6/77.8  & 48.8/72.8  \\ 
    \shortcite{DBLP:journals/taslp/GuoYXY23} & ResNet & CFM+GF+MU & \ding{55} & 42.0/59.4 & 34.1/52.5  & 30.4/49.6  & 62.4/77.2  & 54.1/72.1  & 46.5/66.7  \\ 
    \shortcite{DBLP:conf/emnlp/GuoF0023} & CLIP & SI & \ding{51} & 42.5/-& 36.0/-  & 32.0/-  & 63.7/-  & 56.2/-  & 46.4/-  \\ 
    \shortcite{DBLP:conf/emnlp/ZuoLLZXZ23} & MAE & ML & \ding{55} & 42.6/69.0 & 34.6/62.0  & 31.0/57.2  & 62.2/81.8  & 54.9/76.5  & 45.8/71.2  \\ 
    \shortcite{DBLP:conf/cikm/ChengZLLZ23} & ResNet & CT & \ding{55} & 44.3/69.9 & 37.4/63.7  & 34.3/60.1  & 64.9/83.3  & 57.5/78.5  & 49.4/73.7  \\ 
    \shortcite{DBLP:conf/emnlp/FuFHHW0L23} & CLIP & CFM+MU+UL & \ding{55} & 33.1/-  & 29.1/-  & 24.8/- & 54.1/-& 47.0/-  & 44.4/- \\ 
    \bottomrule
 \end{tabular}
 }
 \end{threeparttable}
\caption{Experimental results of Transformer-based scene-image MMT models. The meanings of symbols in the Techniques column are as follows: adversarial training (AT), back translation (BT), cross-modal filter mechanism (CFM), capsule network (CN), contrastive learning (CT), double attention (DA), gated fusion (GF), graph neural network (GNN), imagination (IG), knowledge distillation (KD), image retrieval (IR), multiple decoding (MD), multitask learning (ML), mix-up (MU), pre-training (PT), reinforcement learning (RL), synthetic images (SI) and unsupervised learning (UL). B/M refers to BLEU/METEOR.} 
\label{tab:SMT}
\end{sidewaystable}


\begin{table}[ht]
\scriptsize
\centering
    \begin{tabular}{l|c|c|c|c|c}
    \toprule
    \textbf{Model}& \textbf{Dataset} & \textbf{Video Encoder} & \textbf{Technique} & \textbf{EN$\Rightarrow$ZH} & \textbf{ZH$\Rightarrow$EN} \\ \midrule
    \rule{0pt}{0ex}
    \shortcite{DBLP:conf/iccv/WangWCLWW19}  & VaTex & 3D ConvNet  & DA & 29.1   & 26.4   \\
    [2pt] \hline \rule{0pt}{3ex}
    \shortcite{DBLP:journals/corr/abs-2006-12799}   & VaTex & ResNet& DA+HAN  & 35.4 & - \\
    [2pt] \hline \rule{0pt}{3ex}
    \shortcite{DBLP:conf/acl/GuSCK21}    & VaTex & 3D ConvNet+Faster R-CNN & DA+HAN  & 35.9   & - \\
    [2pt] \hline \rule{0pt}{3ex}
    \multirow{2}{*}{\shortcite{DBLP:conf/acl/KangHP0SCWHS23}} & VaTex & SlowFast    & \multirow{2}{*}{CT} & 37.6   & -    \\ 
& BigVideo    & Vision Transformer   &  & 44.8   & -    \\
    [2pt] \hline \rule{0pt}{3ex}
    \shortcite{DBLP:journals/pami/LiHCHYH23}    & VaTex & ResNet+Faster-RCNN & ML+PT+UL   & 27.3   & 24.3\\
    [2pt] \hline \rule{0pt}{3ex}
    \shortcite{DBLP:conf/emnlp/LiSCKL23}    & EVA & CLIP4Clip & CFM+GF+ML   & -   & 27.6\\ \bottomrule
    \end{tabular}
\caption{The BLEU scores of video-guided MT models. The HAN technique refers to the hierarchical attention network.}
\label{tab:VMT}
\end{table}


\begin{table}[ht]
\scriptsize
\centering
\setlength{\tabcolsep}{0.45mm}{
    \begin{tabular}{l|c|c|c|c|c}
    \toprule
    \textbf{Model}  & \textbf{Image Encoder}  & \textbf{Technique}& \textbf{Decoding Strategy} & \textbf{EN$\Rightarrow$DE} & \textbf{EN$\Rightarrow$FR} \\ \midrule \rule{0pt}{0ex}
    \multirow{2}{*}{\shortcite{DBLP:conf/wmt/ImankulovaKHK20}} & \multirow{2}{*}{Resnet} & \multirow{2}{*}{HAN}   & consecutive & 34.8   & 53.8  \\
  &   &  & wait-1/2/3& 19.9/-/28.8 & 32.5/-/44.0  \\ [2pt]
    \hline \rule{0pt}{3ex}
    \multirow{2}{*}{\shortcite{DBLP:conf/emnlp/CaglayanIHMBS20}} & \multirow{2}{*}{Faster R-CNN+Resnet}  & \multirow{2}{*}{RL}    & consecutive & 35.3   & 58.1   \\
  &   &  & wait-1/2/3 & 21.3/28.1/32.2 & 42.1/49.2/54.8   \\ [2pt]
    \hline \rule{0pt}{3ex}
    \multirow{2}{*}{\shortcite{DBLP:conf/eacl/IveLMCMS21}} & \multirow{2}{*}{Faster R-CNN+Resnet}  & \multirow{2}{*}{ML+RL} & consecutive & 35.9   & 59.1   \\
  &   &  & wait-1/2/3& -/28.3/32.6   & -/48.1/54.0   \\ [2pt]
    \hline \rule{0pt}{3ex}
    \multirow{2}{*}{\shortcite{DBLP:journals/jair/HaralampievaCS22}}  & \multirow{2}{*}{Faster R-CNN+Resnet} & \multirow{2}{*}{ML}    & consecutive & - & -  \\
  &   &  & wait-1/2/3& 21.4/28.0/31.3 & 42.9/51.5/56.2   \\
    \bottomrule
    \end{tabular}
    }
\caption{The BLEU scores of multi-modal SiMT models in the Test2016 test set of Multi30K. When using the \emph{wait-k} decoding, the decoder waits \emph{k} words to be read before committing the next translation. While using the \emph{consecutive} decoding, the decoder performs translation without waiting.}
\label{tab:SiMT}
\end{table}

\section{Future Directions}
To summarize, previous studies not only employ various techniques to improve translation quality, but also prompt diverse applications of MMT, which have greatly contributed to the development of this task. In our opinion, the future directions of MMT include the following aspects.

\begin{itemize}
\item  \textbf{Visual Information Integration in LLMs.} Very recently, with the rapid development of LLMs \shortcite{DBLP:conf/nips/Ouyang0JAWMZASR22,DBLP:journals/corr/abs-2303-08774,DBLP:journals/corr/abs-2302-13971}, conventional text-only MT has evolved into LLM-based text-only MT. Likewise, utilizing LLM for MMT becomes one future direction in this task. However, existing multi-modal LLMs simply use a linear model \shortcite{DBLP:journals/corr/abs-2304-08485} or Query Transformer \shortcite{DBLP:conf/icml/0008LSH23} to project the visual representations into the semantic space of the textual modality, which may lead to significant loss of visual information. Thus how to enhance the alignments between textual and visual modalities is worth exploring. Besides, existing studies usually apply single-pass methods to extract visual information without any filtering, which will introduce visual noise into the LLMs. How to adaptively extract useful visual information greatly impacts the performance of multi-modal LLMs in the MMT task.
\item  \textbf{Tailored Task Evaluation.} Presently, most existing evaluation metrics for MMT rely on text-only MT metrics, such as BLEU and METEOR. These metrics only consider the similarity between translation and reference within the textual modality, disregarding the semantic information provided by the visual modalities. As a consequence, there is a pressing need for the development of automatic evaluation metrics that concentrate on the semantic matching between translations (especially the ambiguous words) and given images/videos. In addition, more and more studies are beginning to apply LLMs for task evaluation due to their strong semantic comprehension capabilities \shortcite{DBLP:conf/acl/KamallooDCR23,DBLP:conf/acl/WadhwaAW23,DBLP:conf/acl/ChiangL23}. Therefore, how to use LLMs for more comprehensive automatic evaluation of MMT is also one of the future research directions.
\item \textbf{More Extensions.} The conventional  MMT mainly focuses on translating a single sentence along with a single image. Along the line of the development of conventional text-only MT, MMT can also be extended to various multi-modal scenarios, such as multilingual MMT, translating blogs containing both textual content and accompanying images or videos posted on social media, educational materials with texts and illustrations. All of them have wide applications in daily life. 
\item \textbf{High-quality Datasets.} Compared with the datasets used in text-only MT, the commonly used MMT datasets are scale-limited and cover few domains. Therefore, how to efficiently construct high-quality datasets at a larger scale and across multiple domains is also a hot spot for future research. Moreover, in the era of LLMs, instruction-tuning datasets play a pivotal role in fine-tuning LLMs, so it is of vital importance to construct high-quality instruction-tuning datasets for LLM-based MMT.

\end{itemize}

\section{Conclusion}
In this paper, we have presented a comprehensive overview of studies on MMT. First, we describe the methods used in scene-image MMT and other types of MMT. Then we provide detailed information about the datasets used in this task, and introduce some commonly used evaluation metrics. Subsequently, we carefully present and analyze experimental results from representative studies on the unified datasets, observing which techniques perform relatively better. Finally, we discuss possible future research directions in this task. We hope that this survey will serve as a valuable resource for researchers interested in MMT.

\vskip 0.2in
\bibliography{refs_filtered}
\bibliographystyle{theapa}

\end{document}